\documentclass[acmsmall]{acmart}
\pdfoutput=1

\usepackage{times}
\usepackage{epsfig}
\usepackage{graphicx}
\usepackage{amsmath}
\usepackage{bm}
\usepackage{color}
\usepackage{algorithm}
\usepackage{algpseudocode}
\usepackage{float}
\usepackage{bbm}
\usepackage{multirow}
\usepackage{diagbox}
\usepackage{enumitem}
\usepackage{rotating}
\usepackage{float}
\usepackage[export]{adjustbox}
\usepackage{subfig}
\usepackage{booktabs, makecell, tabularx,siunitx}
\usepackage{mathtools}
\DeclarePairedDelimiter{\norm}{\lVert}{\rVert}

\DeclareMathOperator*{\argmax}{arg\,max}

\makeatletter
\def\thickhline{%
  \noalign{\ifnum0=`}\fi\hrule \@height \thickarrayrulewidth \futurelet
   \reserved@a\@xthickhline}
\def\@xthickhline{\ifx\reserved@a\thickhline
               \vskip\doublerulesep
               \vskip-\thickarrayrulewidth
             \fi
      \ifnum0=`{\fi}}
\makeatother

\newlength{\thickarrayrulewidth}
\AtBeginDocument{%
  \providecommand\BibTeX{{%
    \normalfont B\kern-0.5em{\scshape i\kern-0.25em b}\kern-0.8em\TeX}}}

\setcopyright{acmlicensed}
\acmJournal{JETC}
\acmYear{2021}
\acmYear{2020} \acmVolume{1} \acmNumber{1} \acmArticle{1} \acmMonth{1} \acmPrice{15.00}

\acmSubmissionID{JETC-2020-0239}

\begin{document}

\title{Direction-Aggregated Attack for Transferable Adversarial Examples}


\author{Tianjin Huang}
\affiliation{%
  \institution{Eindhoven University of Technology}
  \country{Eindhoven, the Netherlands} 
}
\email{t.huang@tue.nl}

\author{Vlado Menkovski}
\affiliation{%
  \institution{Eindhoven University of Technology}
  \country{Eindhoven, the Netherlands} 
}
\email{v.menkovski@tue.nl}

\author{Yulong Pei}
\affiliation{%
  \institution{Eindhoven University of Technology}
  \country{Eindhoven, the Netherlands} 
}
\email{y.pei.1@tue.nl}

\author{YuHao Wang}
\affiliation{%
  \institution{National University of Singapore}
  \country{Singapore} 
}
\email{yohanna.wang0924@gmail.com}

\author{Mykola Pechenizkiy}
\affiliation{%
  \institution{Eindhoven University of Technology}
  \country{Eindhoven, the Netherlands} 
}
\email{m.pechenizkiy@tue.nl}

\renewcommand{\shortauthors}{T. Huang}

\begin{abstract}
Deep neural networks are vulnerable to adversarial examples that are crafted by imposing imperceptible changes to the inputs. 
However, these adversarial examples are most successful in white-box settings where the model and its parameters are available. Finding adversarial examples that are transferable to other models or developed in a black-box setting is significantly more difficult. 
In this paper, we propose the Direction-Aggregated adversarial attacks that deliver transferable adversarial examples. Our method utilizes the aggregated direction during the attack process for avoiding the generated adversarial examples overfitting to the white-box model.
Extensive experiments on ImageNet show that our proposed method improves the transferability of adversarial examples significantly and outperforms state-of-the-art attacks, especially against adversarial trained models. The best averaged attack success rate of our proposed method reaches 94.6\% against three adversarial trained models and 94.8\% against five defense methods. It also reveals that current defense approaches do not prevent transferable adversarial attacks.
\end{abstract}




\maketitle

\section{Introduction}
\label{Introduction}
Deep Neural Networks (DNNs) have achieved a great success in many tasks, e.g. image classification~\cite{Krizhevsky2012,He2016}, object detection~\cite{Girshick2014}, segmentation~\cite{Long2015}, etc. However, these high-performing models have been shown to be vulnerable to adversarial examples~\cite{Szegedy2013,GoodfellowExplaining}. In other words, carefully crafted changes to the inputs can change the model's prediction drastically. This fragility has raised concerns on security-sensitive tasks such as autonomous cars, face recognition, and malware detection. Well designed adversarial examples are not only useful to evaluate the robustness of models against adversarial attacks but also beneficial to improve the robustness of them~\cite{GoodfellowExplaining}. 

Plenty of ways have been proposed to craft adversarial examples, which can be divided into white-box and black-box attacks. White-box attacks utilize complete knowledge including model architecture, model parameters, training strategy and training method, e.g. fast gradient sign method (FGSM)~\cite{GoodfellowExplaining}, Iterative Fast Gradient Sign Method (I-FGSM)~\cite{kurakin2016adversarial}, Project gradient descent (PGD)~\cite{Madry2017}, Deepfool~\cite{Moosavi-Dezfooli2016}, Momentum Iterative Fast Gradient Sign Method (MI-FGSM)~\cite{dong2018boosting} and Carlini \& Wagner's attack~\cite{carlini2017towards}. On the contrary, black-box attacks fool the model's prediction without any knowledge about the model. It has been shown that adversarial examples generated by white-box attacks have the ability to fool other black-box models, which is known as the transferability property~\cite{Szegedy2013}. The transferability of adversarial examples enables practical black-box attacks and imposes a huge threat on real-world applications. However, the transferability of adversarial examples usually is very low because these adversarial examples easily overfit to the white-box model, i.e. the model for generating these adversarial examples. Therefore, avoiding the \emph{overfitting} problem is the key to generate transferable adversarial examples. 

\begin{figure}[htb]
    \centering
        \includegraphics[width=0.5\textwidth]{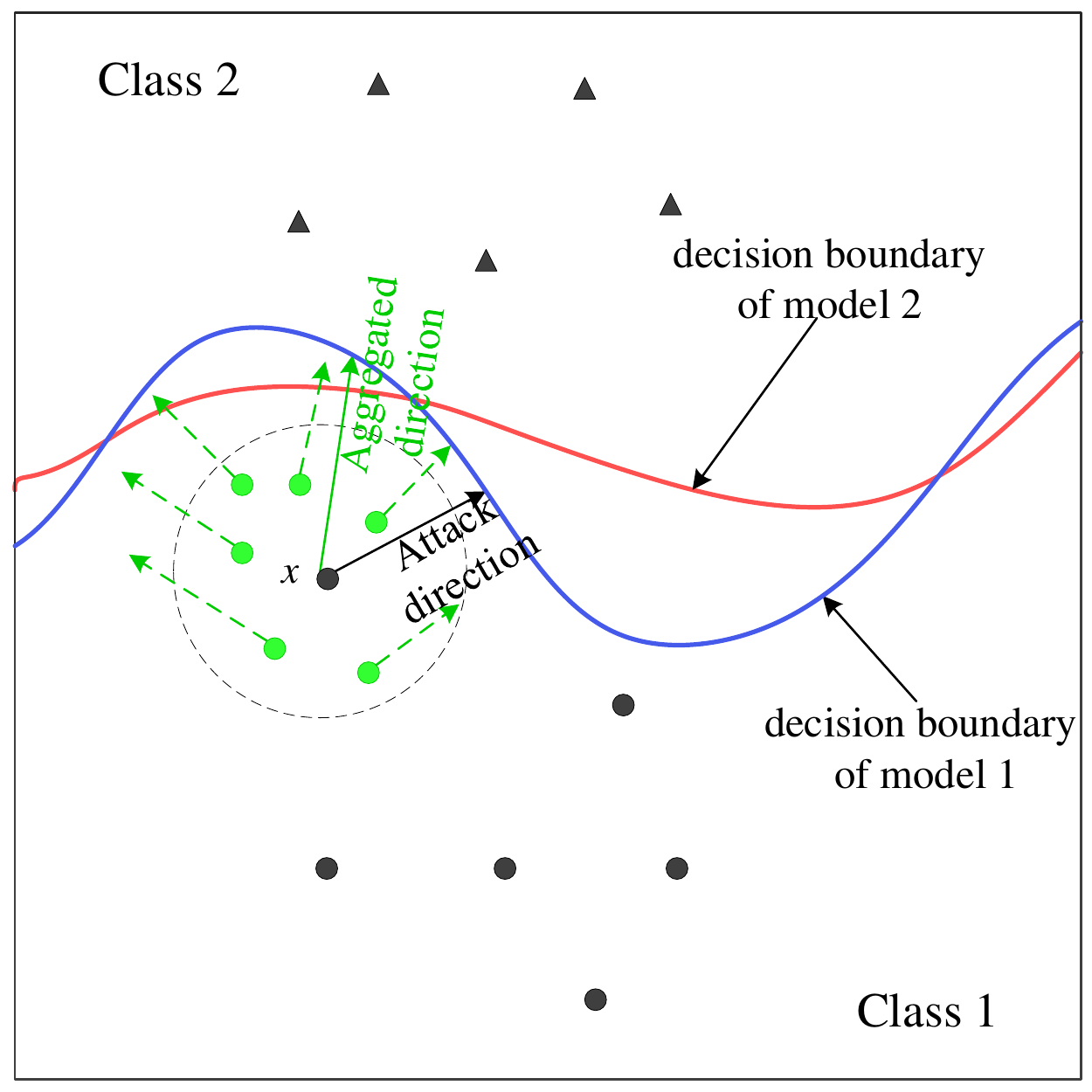}
    \caption{A simple schematic diagram for explaining why aggregated direction can mitigate the overfitting problem of adversarial examples. Black circle and triangle markers denote samples of class 1 and class 2 respectively. Red and blue lines represent the decision boundary of model 1 and model 2. Circle with dotted line denotes a set of examples from the neighborhood of $x$ . Black arrow line denotes the attack direction ($sgn(\nabla_{\bm{x}} L(f_{\theta}(\bm{x}),y))$) of model 1 at the sample $x$. Green arrow dotted lines are the attack direction at the perturbed sample with Gaussian noise. Green arrow solid lines denote the aggregated direction by the vector addition of the green arrow dotted lines.}.  
    \label{fig:1}
\end{figure}

 Deep neural networks applied to high dimensional classification tasks are typically very complex models, in other words, the decision boundary is highly non-linear and tends to have high curvature, e.g., the decision boundary of \textsl{model 1} in Fig.~\ref{fig:1}. We believe that it is the high curvature of a decision boundary that makes adversarial examples decrease their ability to attack other models especially adversarial robust models~\footnote{In this paper, it denotes a model trained with an adversarial training technique.} that have smoothed decision boundary~\cite{cohen2019certified,li2018certified}. As shown in Fig.~\ref{fig:1}, the adversarial attack direction generated by model 1 at sample $x$ (the black arrow line in Fig.~\ref{fig:1}) tends to overfit to model 1 because this attack direction is the best direction~\footnote{It denotes the direction that is perpendicular to the decision boundary.} for attacking \textsl{model 1}, but not a good direction for attacking \textsl{model 2}. To mitigate the issue of adversarial examples easily overfitting to the white-box model, we propose to aggregate the attack directions from the neighborhood of the input $x$, e.g., by adding Gaussian noise or Uniform noise to the input. The green solid arrow line in Fig.~\ref{fig:1} shows the aggregated direction. It is easy to see that the green solid arrow line is a good attack direction for both \textsl{model 1} and \textsl{model 2}. Therefore, adversarial examples generated by the aggregated direction can achieve good transferability. Based on this, we propose the Direction-Aggregated attack (DA-Attack) for improving the transferability of adversarial examples. Results of the extensive experiments presented in later sections show that our method achieves state-of-the-art results. 

In detail, our contributions are summarized as follows:
\begin{itemize}
    \item We propose to aggregate attack directions in order to stabilize the oscillation of attack directions and guide the attack direction to the generalized decision boundary and avoid overfitting to the white-box model's decision boundary. Based on the aggregated direction, we propose our DA-Attack.
    \item We demonstrate experimentally that DA-Attack outperforms state-of-the-art attacks by extensive experiments on ImageNet. The best averaged attack success rate of our method achieves 94.6\% against three ensemble adversarial trained models and 94.8\% against five defense methods, which also reveals that current defense models are not safe to transferable adversarial attacks. We expect that the proposed DA-Attack will serve as a benchmark for evaluating the effectiveness of adversarial defense methods in the future.  
    \item We experimentally show that sampling times $N$, standard deviation $\sigma$, iterations $T$ and perturbation size $\epsilon$ induced in our method play an important role in achieving the transferability of adversarial examples. Usually a bigger value in $N$, $\sigma$, $T$ and $\epsilon$ can lead to a higher transferability of the adversarial examples. However, a too large value in $T$ and $\sigma$ would lead to a negative effect.
\end{itemize}

The rest of this paper is organized as follows. In Section~\ref{rw} we present related work. In Section~\ref{method} we describes our proposed DA-Attack in detail. In Section~\ref{exp} we discuss the results of the extensive experiments with DA-Attack. In Section~\ref{dis} we discuss the connection of the DA-Attack to a smoothed classifier. We draw conclusions in Section~\ref{conclude}.

\section{Related Work}\label{rw}
\textbf{Adversarial examples}
Szegedy et al.~\cite{Szegedy2013} first found the existence of adversarial examples: given an input $(x,y)$ and a classifier $f_{\theta}$, it is possible to find a similar input $x^*$ such that $f_{\theta}(x^*) \neq y$. A formal mathematical definition is as follows:
\begin{align}
    \min_{x^*} \norm{x^*-x}_{p},
    \;\; s.t. \;\; f_{\theta}(x^*)\neq y,\;
         f_{\theta}(x)=y
\end{align}
where $\norm{\cdot}_{p}$ denotes the $L_{p}$ distance.

Following~\cite{Szegedy2013}, many related researches have emerged. On the one hand, some of them propose to generate adversarial examples that can be applied in the physical world~\cite{Elsayed2018,kurakin2016adversarial}. On the other hand, some of them focus on reducing the minimal size of adversarial perturbations and improving the attack success rates~\cite{GoodfellowExplaining,dong2018boosting,carlini2017towards,Moosavi-Dezfooli2016}. Among these researches, the attack success rates under the black-box setting is still low, especially against adversarial trained models, i.e. the model is trained by adversarial training technique which can effectively defend against adversarial examples~\cite{Madry2017}. Recently, several papers improve the attack success rates based on transferable adversarial attacks. 
Inkawhich et al.~\cite{inkawhich2019feature} generate more transferable adversarial examples by enlarging the distance between adversarial examples and clean samples in feature space. Their intuition is from the fact that deep feature representations of models are transferable. Similar in utilizing feature representations, Zhou et al.~\cite{zhou2018transferable} improve the transferability by reducing the variations of adversarial perturbations via constructing a new regularization based on feature representations.
Liu et al.~\cite{liu2016delving} demonstrate that the transferability can be improved by attacking an ensemble of substitute models. This method suffer from expensive computational cost since multiple models are needed to be trained first. Li et al.~\cite{li2018learning} further reduce the computation cost of the method by attacking ``Ghost Networks'' where the ``Ghost Networks'' are generated from a basic trained model.  Xie et al.~\cite{xie2019improving} believe that overfitting to the white-box model decreases the transferability of adversarial examples, therefore they induce the data augmentation technique to mitigate the overfitting issue. Specifically, they apply random transformations to the inputs and calculate gradient based on the transformed inputs. Dong et al.~\cite{dong2019evading} find that different models make predictions based on different discriminative regions of the input, which decreases the transferability of adversarial examples. Based on this intuition, they propose a translation-invariant attack by averaging the gradients from an ensemble of images composed of the image and its translated versions. Similarly, Lin et al.~\cite{Lin2020Nesterov} enhance the transferability of adversarial examples by averaging gradients from an ensemble of images composed of the image and its scaled versions. Besides, Lin et al.~\cite{Lin2020Nesterov} also demonstrate that Nesterov accelerated gradient can further improve the transferability of adversarial examples. Wu and Zhu~\cite{wu2020towards} improve the transferability of adversarial examples by smoothing the loss surface. Our method is degraded to this method when the attack direction of each step is the gradient of loss w.r.t the inputs. Naseer et al.~\cite{Naseer2019Cross} propose ``domain-agnostic'' adversarial perturbations which can be used to fool models learned from different domains. 
\\
\textbf{Defense against adversarial examples}
Correspondingly, many methods have been proposed to defend against these adversarial examples. Usually, the ability of a model for defending adversarial examples is referred to adversarial robustness. It measures a model's resilience against adversarial examples. Goodfellow, Shlen and Szegedy~\cite{GoodfellowExplaining}, Madry et al.~\cite{Madry2017} effectively improve a model's adversarial robustness by adversarial training technique. That is, it trains model based on on-the-fly generated adversarial examples $x^{*}$ bounded by uniformly $\epsilon$-ball of the input x (i.e., $\norm{x^{*}-x} \leq \epsilon$). Tramer et al.~\cite{tramer2017ensemble} further improve adversarial robustness by ensemble adversarial training where the model is trained on adversarial examples generated from multiple pretrained models. Cohen, Rosenfeld and Kolter~\cite{cohen2019certified} build guaranteed adversarial robust model by transforming a base classifier $f$ into a smoothed classifier's $g$. Specifically, the prediction of $g(X)$ is defined to be the class which $f$ is most likely to classify the random variable $\mathcal{N}(x,\sigma^{2}I)$ as.
On the other hand, several papers try to defend against adversarial examples by purifying or reducing adversarial perturbations.
 Xie et al.~\cite{xie2017mitigating} and Guo et al.~\cite{guo2017countering} impose transformations, e.g., image cropping, rescaling, quilting, padding and so on, on input images at inference time to reduce the adversarial perturbations, and therefore increase accuracy of the model's performance on adversarial examples. Liao et al.~\cite{liao2018defense} propose a U-net based denoiser to purify the adversarial perturbations.  

\section{Methodology}\label{method}
In this section, we first introduce notation and then provide details of our method.

\subsection{Notation}
We specify the notations that are used in this paper by the following list:
\begin{itemize}
    \item $\bm{x}$ and $y$ denote a clean image and the corresponding true label respectively.
    \item $\bm{x}^*$ denotes the adversarial example.
    \item $f_{\theta}(\bm{x})$ denotes a deep neural network.
    \item $L(f_{\theta}(\bm{x}),y)$ represents the Cross-Entropy loss.
    \item $sgn(\cdot)$ denotes the sign function.
    \item $\nabla_{\bm{x}}L(\cdot)$ denotes the gradient of $L(\cdot)$ with respect to $\bm{x}$.
    \item $Clip_{\bm{x}}^{\epsilon}(\cdot)$ function limits the generated adversarial example $\bm{x}^{*}$ to the $\epsilon$ max-norm ball of $\bm{x}$.
    \item $\epsilon$ is the allowed maximum perturbation size of the adversarial perturbation.
    \item $\alpha$ is the step size for PGD/FGSM-based adversarial attacks.
    \item $\mathcal{N}(0,\sigma^{2}I)$ denotes Gaussian distribution with mean 0 and standard deviation $\sigma$.
    \item $U(a,b)$ is Uniform distribution.
    \item $\varepsilon$ denotes a small random noise and can be generated from Gaussian distribution or Uniform distribution. In this paper, we adopt Gaussian noise by default. 
    \item $|\cdot|$ denotes the number of elements of a set.
    \item $D^*$ denotes a set of adversarial examples.
\end{itemize}

\subsection{Gradient-based Adversarial Attack Methods}
Several adversarial attacks will be integrated into our proposed method. We give a brief introduction of them in this section.

\textbf{Fast Gradient Sign Method (FGSM)}~\cite{GoodfellowExplaining} generates adversarial examples by adding a fixed magnitude along the sign of gradients of the loss function, which is formalized as follows:
\begin{align}
\bm{x}^{*}=\bm{x}+\epsilon\cdot sgn(\nabla_{\bm{x}} L(f_{\theta}(\bm{x}),y)).
\label{eq:fgsm}
\end{align}

\textbf{Iterative Fast Gradient Sign Method (I-FGSM)}~\cite{kurakin2016adversarialphysical} is a multi-step variant of FGSM and restricts the perturbed size to the $\epsilon$ max-norm ball. With the initialization $\bm{x}_{0}^{*}=x$, the perturbed data in $t-th$ step $\bm{x}_{t}^{*}$ can be expressed as follows:
\begin{equation}
\bm{x}_{t}^{*}=Clip_{\bm{x}}^{\epsilon} \{\bm{x}_{t-1}^{*}+\alpha\cdot sgn(\nabla_{\bm{x}} L(f_{\theta}(\bm{x}_{t-1}^{*}),y))\}.
\end{equation}

\textbf{Momentum iterative fast gradient sign method (MI-FGSM)}~\cite{dong2018boosting} integrates momentum into I-FGSM method for stabilizing optimization, which can be expressed as follows:
\begin{align}
\bm{g}_{t+1}=\mu\cdot\bm{g}_{t}+\frac{\nabla{\bm{x}}L(f_{\theta}(\bm{x}_{t}^{*}),y)}{\norm{\nabla{\bm{x}}L(f_{\theta}(\bm{x}_{t}^{*}),y)}_{1}}  \\
\bm{x}^{*}_{t+1}=Clip_{\bm{x}}^{\epsilon}\{\bm{x}_{t}^{*}+\alpha\cdot sgn(\bm{g}_{t})\}
\end{align}
where $\bm{g}_{t}$ is the accumulated gradient at iteration $t$ and $\mu$ is the decay factor of the momentum term.

\textbf{Diverse Inputs Method(DIM)}~\cite{xie2019improving} calculates gradient based on random transformed inputs. The transformation contains random resizing and padding with a given probability. Formally, it  can be expressed as follows:
\begin{align}
\bm{g}_{t+1}=\mu\cdot\bm{g}_{t}+\frac{\nabla{\bm{x}}L(f_{\theta}(T(\bm{x}_{t}^{*};p)),y)}{\norm{\nabla{\bm{x}}L(f_{\theta}(T(\bm{x}_{t}^{*};p)),y)}_{1}}  \\
\bm{x}^{*}_{t+1}=Clip_{\bm{x}}^{\epsilon}\{\bm{x}_{t}^{*}+\alpha\cdot sgn(\bm{g}_{t+1})\}
\end{align}
where $T(\cdot;p)$ is the stochastic transformation function and $p$ is the transformation probability.

\textbf{Translation-invariant Method(TIM)}~\cite{dong2019evading} generates an adversarial example by an ensemble of translated inputs and it was demonstrated to be equivalent to convolve the gradient at the untranslated image. Specifically, it can be expressed as follows:
\begin{align}
\bm{g}_{t+1}=\mu\cdot\bm{g}_{t}+\frac{\mathcal{W} *\nabla{\bm{x}}L(f_{\theta}(\bm{x}_{t}^{*}),y)}{\norm{\mathcal{W} *\nabla{\bm{x}}L(f_{\theta}(\bm{x}_{t}^{*}),y)}_{1}}  \\
\bm{x}^{*}_{t+1}=Clip_{\bm{x}}^{\epsilon}\{\bm{x}_{t}^{*}+\alpha\cdot sgn(\bm{g}_{t})\}
\end{align}
where $*$ is the convolutional operation and $\mathcal{W}$ is the kernel matrix of size $(2k+1) \times (2k+1)$. Following~\cite{dong2019evading}, a Gaussian kernel is chosen for our experiments. It is defined as: $\mathcal{\Tilde{W}}_{i,j}=\frac{1}{2\pi\sigma^2} \exp{- \frac{i^2+j^2}{2\sigma^2}}$ where the standard deviation $\sigma=k/\sqrt{3}$ and  $\mathcal{W}_{i,j}=\mathcal{\Tilde{W}}_{i,j}/\sum_{i,j} \mathcal{\Tilde{W}}_{i,j}$.

\subsection{Direction-Aggregated Attack (DA-Attack)}
In Fig.~\ref{fig:1} we illustrated that adversarial examples could overfit to the white-box model due to the very complex decision boundary decreasing their transferability. We mitigate this problem with overfitting adversarial examples by aggregating the attack directions of a set of examples from the neighborhood of the input. We integrate the aggregated direction to basic adversarial attacks, i.e.\ Fast Gradient Sign Method (FGSM)~\cite{Szegedy2013}, Iterative Fast Gradient Sign Method (I-FGSM)~\cite{kurakin2016adversarialphysical}, and Momentum Iterative Fast Gradient Sign Method (MI-FGSM)~\cite{dong2018boosting}, for improving their transferability. Besides, to further enhance the transferability, we combine our method with other transferable adversarial attacks, i.e. Diverse Input Method (DIM)~\cite{xie2019improving}, Translation-Invariant Method (TIM)~\cite{dong2019evading}, TI-DIM~\cite{dong2019evading}. Concretely, the update procedures for each attack are formalized as follows.

\textbf{DA-FGSM.}
To mitigate the effect of overfitting to the specific model and improve the transferability of adversarial examples for FGSM attack, we propose the Direction-Aggregated FGSM (DA-FGSM). 
The attack direction is replaced with the aggregated direction which are achieved by aggregating the attack directions of a set of examples from the neighborhood of the input $x$. In practice, we generate the set of examples by adding small perturbations to the input, i.e. adding Gaussian noise or Uniform noise to the input. In this paper, we adopt Gaussian noise as the default choice. We further provide the evidence that Uniform noise can reach the same performance as Gaussian noise. Formally, it can be represented as follows:
\begin{align}
\bm{x}^{*}=\bm{x}+\epsilon\cdot sgn(\sum_{i=0}^{N}(sgn(\nabla_{\bm{x}} L(f_{\theta}(\bm{x}+\bm{\varepsilon}_i),y)))),
\label{eq:2}
\end{align}
where $N$ denotes the sampling times from certain noise distribution. The $sgn(\nabla_{\bm{x}} L(f_{\theta}(\bm{x}+\bm{\varepsilon}_i),y))$ denotes one specific attack direction. We aggregate the $N$ attack directions by the sum operation. 

\textbf{DA-I-FGSM.}
To improve the transferability for I-FGSM. We propose the Direction-Aggregated I-FGSM (DA-I-FGSM). 
The attack direction at each iteration is replaced with the aggregated direction. The update procedure can be formalized as follows:
\begin{equation}
\bm{x}_{t}^{*}=Clip_{\bm{x}}^{\epsilon} \{\bm{x}_{t-1}^{*}+\alpha\cdot sgn(\sum_{i=0}^{N}(sgn(\nabla_{\bm{x}} L(f_{\theta}(\bm{x}_{t-1}^{*}+\bm{\varepsilon}_i),y))))\}.
\label{eq:4}
\end{equation}

\textbf{DA-MI-FGSM.}
We integrate the momentum term into DA-I-FGSM for improving the attack ability, which is called Momentum Direction-Aggregated I-FGSM (DA-MI-FGSM). The update procedure of DA-MI-FGSM can be expressed as follows:
\begin{align}
\bm{g}_{a}&=\sum_{i=0}^{N}(sgn(\nabla_{\bm{x}} L(f_{\theta}(\bm{x}_{t-1}^*+\bm{\varepsilon}_i),y)))  \label{eq:5}\\
\bm{g}_t&=\mu \cdot \bm{g}_{t-1}+\frac{\bm{g}_{a}}{\norm{\bm{g}_{a}}_1} \label{eq:6}\\
\bm{x}_{t}^{*}&=Clip_{\bm{x}}^{\epsilon} \{\bm{x}_{t-1}^{*}+\alpha\cdot sgn(\bm{g}_{t})\}, 
\label{eq:7}
\end{align}
where $\bm{g}_t$ is the accumulated gradient at iteration $t$ and $\mu$ is the decay factor of the momentum term, and $\bm{g}_a$ is the aggregated direction. 

\textbf{DA-DIM.}
We combine our proposed DA-MI-FGSM with DIM for further improving the transferability of adversarial examples and denote it as Direction-Aggregated DIM (DA-DIM). The update procedure is similar to DA-MI-FGSM, with the replacement of Eq. (\ref{eq:5}) by the following equation:
\begin{equation}
    \bm{g}_{a}=\sum_{i=0}^{N}(sgn(\nabla_{\bm{x}} L(f_{\theta}(T(\bm{x}_{t-1}^*+\bm{\varepsilon}_i;p)),y))),
    \label{eq:8}
\end{equation}

\textbf{DA-TIM.}
Similar to DA-DIM, we combine DA-MI-FGSM with TIM and denote it as Direction-Aggregated TIM (DA-TIM). Likewise, the update procedure is similar to DA-MI-FGSM, with the replacement of Eq. (\ref{eq:6}) by the following equation:
\begin{align}
    \bm{g}_t=\mu\cdot \bm{g}_{t-1}+\frac{\mathcal{W} * \bm{g}_a}{\norm{\mathcal{W} * \bm{g}_a}_1},
    \label{eq:9}
\end{align}

\textbf{DA-TI-DIM.}
Following~\cite{Lin2020Nesterov}, we combine DA-MI-FGSM with TIM and DIM together and denote it as Direction-Aggregated TI-DIM (DA-TI-DIM). The update procedure can be presented as follows:
\begin{align}
\bm{g}_{a}&=\sum_{i=0}^{N}(sgn(\nabla_{\bm{x}} L(f_{\theta}(T(\bm{x}_{t-1}^*+\bm{\varepsilon}_i;p)),y))) \\
\bm{g}_t&=\mu\cdot \bm{g}_{t-1}+\frac{\mathcal{W} * \bm{g}_a}{\norm{\mathcal{W} * \bm{g}_a}_1}\\
\bm{x}_{t}^{*}&=Clip_{\bm{x}}^{\epsilon} \{\bm{x}_{t-1}^{*}+\alpha\cdot sgn(\bm{g}_{t})\}. 
\label{eq:10}
\end{align}

\begin{algorithm}
\caption{DA-MI-FGSM}
\label{alg:1}
\begin{algorithmic}[1]
\Require
A input image $x$ with true label $y$; a classifier $f$ with loss function $L$; perturbation size $\epsilon$; maximum iterations $T$; Gaussian distribution $\mathcal{N}(0,\sigma^{2}I)$; The decay factor $\mu$; the aggregated direction $\bm{g}_a$.
\Ensure
An adversarial example $\bm{x}^*$
\State $\alpha$ = $\epsilon$/$T$
\State $\bm{x}_0^{*}$=$\bm{x}$; $\bm{g}_0=0$
\For{$t=1$ to $T$}
    \State $\bm{g_a}$=0
    \For{$i=0$ to $N$}  
      \State Get $\bm{\varepsilon}_i$ $\sim$ $\mathcal{N}(0,\sigma^2\mathcal{I})$
      \State Aggregate attack directions as $\bm{g}_a$=$\bm{g}_a$+$sgn(\nabla_{\bm{x}} L(f_{\theta}(\bm{x}_{t-1}^{*}+\bm{\varepsilon}_i),y))$
    \EndFor
    \State Update $\bm{g}_{t}$=$\mu \cdot \bm{g}_{t-1}$+$\frac{\bm{g}_{a}}{\norm{\bm{g}_{a}}_1}$
    \State Update $\bm{x}_t^*=Clip_x^{\epsilon}\{\bm{x}_{t-1}^*+\alpha \cdot sgn(\bm{g}_t) \}$
\EndFor
\State $\bm{x}^*=\bm{x}^*_t$ \\
\Return $\bm{x}^*$
\end{algorithmic}
\end{algorithm}

The pseudocode of DA-MI-FGSM is summarized in Algorithm~\ref{alg:1} and the code is provided\footnote{\url{https://github.com/Juintin/DA-Attack.git}}.

\section{Experiments}\label{exp}
We evaluate the effectiveness of DA-Attack empirically. We first introduce the dataset and experimental settings. Then we show the performance of our method against normal and defense models. Finally, we analyze the influence of the parameters $N$, $\sigma$, $\epsilon$, $T$ and $\alpha$ on achieving the transferability of adversarial examples. 
\subsection{Experimental Settings}

\textbf{Datasets.}  Following the strategy used in~\cite{Lin2020Nesterov}, a set of 1000 images (denoted as $D$) that are correctly classified by all testing models are randomly selected  from ILSVRC 2012 validation set. For a fair comparison with state-of-the-art methods, we use the same 1000 images\footnote{\url{https://github.com/JHL-HUST/SI-NI-FGSM}} in~\cite{Lin2020Nesterov}.\\
\textbf{Models.}
Four normal trained models and three ensemble adversarial trained models are used for evaluating adversarial examples, which are Inception-V3 (Inc-V3)~\cite{szegedy2016rethinking}, Inception-v4 (Inc-V4)~\cite{szegedy2017inception}, Inception-Resnet-v2 (IncRes-V2)~\cite{szegedy2017inception}, Resnet-V2 (Res-101)~\cite{He2016}, Inc-V3$_{ens3}$, Inc-V3$_{ens4}$ and IncRes-V2$_{ens}$~\cite{tramer2017ensemble} respectively. Besides, five advanced defense methods are considered for further evaluating the effectiveness of our method. Specifically, the selected advanced defense methods are High-level representation guided denoiser (HGD)~\cite{liao2018defense}, Random resizing and padding (R\&P)~\cite{xie2017mitigating}, NIPS-r3\footnote{\url{https://github.com/anlthms/nips-2017/tree/master/mmd}}, feature distillation (FD)~\cite{liu2019feature} and purifying perturbations by image compression (Comdefend)~\cite{jia2019comdefend}.\\
\textbf{Baselines.}
Several most recently proposed methods aiming at generating transferable adversarial examples are taken as baselines:
\begin{itemize}
    \item DIM~\cite{xie2019improving}, which generates transferable examples by random resizing input images;
    \item TIM~\cite{dong2019evading}, which generates transferable examples by a set of translated images;
    \item SI-NI-FGSM~\cite{Lin2020Nesterov}, which generates transferable examples by scaled images and nesterov accelerated gradients; and
    \item  The combinations of DIM, TIM and SI-NI-FGSM, namely TI-DIM~\cite{dong2019evading}, SI-NI-TIM~\cite{Lin2020Nesterov}, SI-NI-DIM~\cite{Lin2020Nesterov} and SI-NI-TI-DIM~\cite{Lin2020Nesterov} attacks.
\end{itemize}
Considering that we completely follow the experimental settings in~\cite{Lin2020Nesterov}, all the baseline results except for the attack success rates against FD and ComDefend in Table~\ref{tab:6} are from~~\cite{Lin2020Nesterov}.\\
\textbf{Hyper-Parameters.}
We follow the settings in~\cite{Lin2020Nesterov} for all hyper-parameters, the maximum perturbation $\epsilon$ is set to $16$ and the number of iterations $T$ is set to $12$ as default values. Accordingly $\alpha = \epsilon/T$. The momentum parameter $\mu$ is set to $1.0$. For DIM and TI-DIM methods, the transformation probability is set to 0.5. For TIM method, Gaussian kernel is adopted as our baseline experiments and kernel size is set to $7\times7$. For SI-NI-FGSM, SI-NI-TIM, SI-NI-DIM and SI-NI-TI-DIM methods, the number of scales is set to 5. For our DA-Attack, sampling times $N$ and standard deviation $\sigma$ are set to 30 and 0.05 respectively.\\
\textbf{Criteria.}
We use the attack success rates to reflect the ability of adversarial examples attacking a model. The attack success rates is defined as follows:
\begin{equation}
    100 \times \frac{\sum_{i=1}^{M} {[\argmax_j f_j(x_i^*) \ne y_i]}}{M}, 
    \label{eq:11}
\end{equation}
where $(x_i^*,y_i) \in D^*$ and $M$ is the number of adversarial examples in $D^*$.

\subsection{Single-Model Attacks}
We first evaluate the effectiveness of DA-Attack based on the single model. DIM~\cite{xie2019improving}, TIM~\cite{dong2019evading} and SI-NI-FGSM~\cite{Lin2020Nesterov} and their combinations, i.e. SI-NI-TIM, TI-DIM, SI-NI-TI-DIM, are taken as baselines. Besides, several popular normal adversarial attacks, i.e. FGSM, I-FGSM, MI-FGSM, PGD, C\&W, are utilized to show the effectiveness of our method.  

\textbf{Comparison with normal and transferable attacks.} The attack success rates of DIM, TIM, SI-NI-FGSM, normal attacks and our proposed method are shown in Table~\ref{tab:1}. The adversarial examples are crafted based on Inc-V3 model. From Table~\ref{tab:1}, it can be observed:
\begin{itemize}
    \item Adversarial examples are much easier to attack normal trained models than adversarial trained models.
    \item Adversarial examples generated by transferable attacks have much higher attack success rates against black-box models than normal attacks.
    \item Our proposed M-ADI-FGSM attack outperforms the current state-of-the-art SI-NI-FGSM attack by 4.6\% to 10.4\%. Besides, DA-FGSM and DA-I-FGSM attacks without momentum acceleration still achieve remarkable results compared with normal attacks, which demonstrate the effectiveness of the aggregated direction.
\end{itemize}
 Besides, it is worthy noting that adversarial examples from I-FGSM attack are less transferable than that from FGSM attack (by comparing I-FGSM with FGSM in Table~\ref{tab:1}), which shows the evidence that adversarial examples overfitting to the white-box model decreases the transferability. And the transferability  is improved by adding a momentum term during generating adversarial examples (by comparing MI-FGSM with I-FGSM in Table~\ref{tab:1}), which conforms the claim in ~\cite{dong2018boosting}. Interestingly, the combination of Direction Aggregation and momentum can greatly improve the transferability again (by comparing MI-FGSM with DA-MI-FGSM in Table~\ref{tab:1}). We conjecture that it is because the proposed Direction Aggregation technique is orthogonal to the momentum technique. Intuitively, Direction Aggregation technique stabilize the attack direction by reducing the oscillation of each update direction during the iterations while momentum stabilize the attack direction by accumulating historical update directions. 

\textbf{Comparison with the extensions of DIM and TIM.} To fully evaluate DA-TIM, DA-DIM and DA-TI-DIM attacks, adversarial examples are crafted by these attacks based on Inc-V3, Inc-V4, IncRes-V2, Res-101 models respectively. We test it against the four normal trained and three ensemble adversarial trained  models. The evaluation results are shown in Table~\ref{tab:2}, Table~\ref{tab:3} and Table~\ref{tab:4}. It can be observed from these results:
\begin{itemize}
    \item The combinations of our method and DIM, TIM methods can greatly improve the transferability of adversarial examples, which indicates that our method is orthogonal to these methods.
    \item Our method outperforms the state-of-the-art attacks across all conducted experiments, i.e.\ SI-NI-TIM, SI-NI-DIM and SI-NI-TI-DIM, except for adversarial examples crafted on IncRes-V2 model. Besides, the attack success rates of our method against the adversarial trained models outperform state-of-the-art attacks by large margins.
\end{itemize}
For the exception that our method does not outperform the state-of-the-art results for adversarial examples crafted on IncRes-V2 model, it may because the adversarial examples generated by our method underfit the IncRes-V2 model somehow since the attack success rates for the white-box model IncRes-V2 is only around 95\% and 4\%-5\% lower than the SI-NI-TIM/DIM method. One possible solution for this ``underfit'' problem is to increase the Iterations T. The results in Fig.~\ref{fig:5c} also indicate that the attack success rates for normal models can be improved a lot by increasing the Iterations T. Besides, we notice that the improvement of combining DA technique and DIM/TIM implemented on different white-box models are different. We think it may be caused by the different degree of non-linearity on the decision boundaries of different white-box models. Intuitively, the higher degree of non-linearity  the decision boundary is, the larger improvements of transferability the DA technique can make. 

\begin{figure}[htb]
 \centering
\setlength\tabcolsep{1pt}
\settowidth\rotheadsize{Radcliffe Cam}
\begin{tabularx}{0.8\linewidth}{l XXXXXX }
\rothead{\centering{Clean}} 
                        &   \includegraphics[width=\hsize,valign=m]{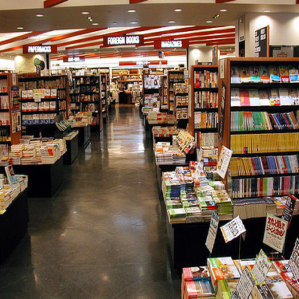}    
                        &   \includegraphics[width=\hsize,valign=m]{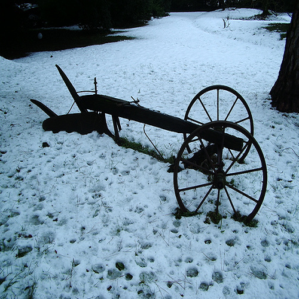}
                        &   \includegraphics[width=\hsize,valign=m]{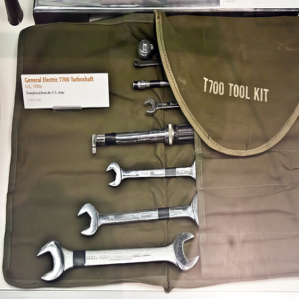}    
                        &   \includegraphics[width=\hsize,valign=m]{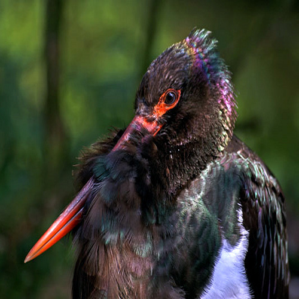}
                        &   \includegraphics[width=\hsize,valign=m]{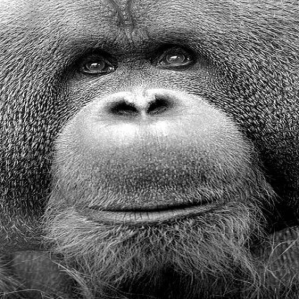}\\[-1.5em]
\rothead{\centering{TIM}} 
                        &   \includegraphics[width=\hsize,valign=m]{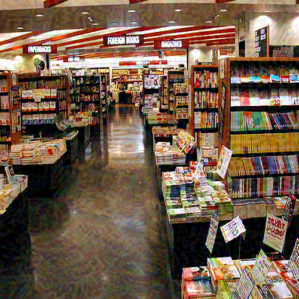}
                        &   \includegraphics[width=\hsize,valign=m]{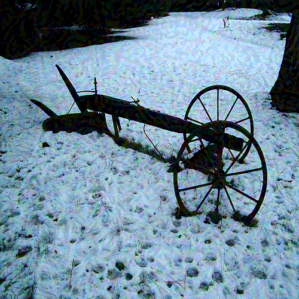}
                        &   \includegraphics[width=\hsize,valign=m]{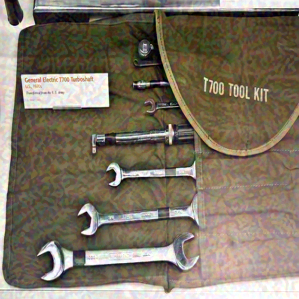}
                        &   \includegraphics[width=\hsize,valign=m]{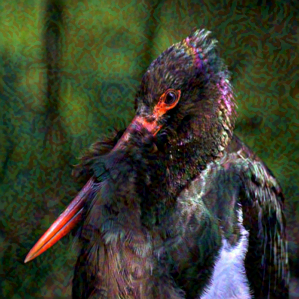}
                        &   \includegraphics[width=\hsize,valign=m]{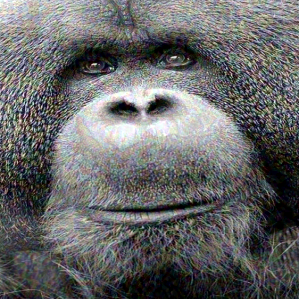}\\[-1.5em]
\rothead{\centering{DIM}} 
                        &   \includegraphics[width=\hsize,valign=m]{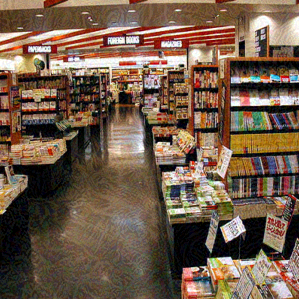}    
                        &   \includegraphics[width=\hsize,valign=m]{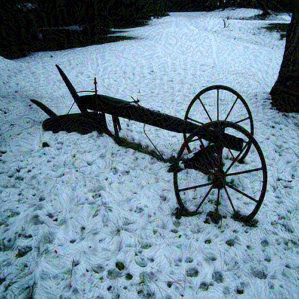}
                        &   \includegraphics[width=\hsize,valign=m]{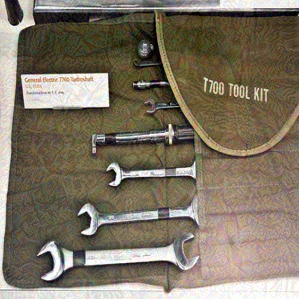}    
                        &   \includegraphics[width=\hsize,valign=m]{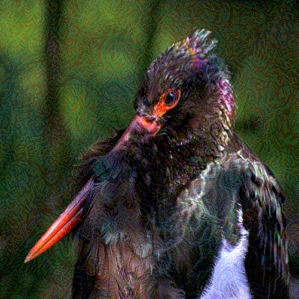}
                        &   \includegraphics[width=\hsize,valign=m]{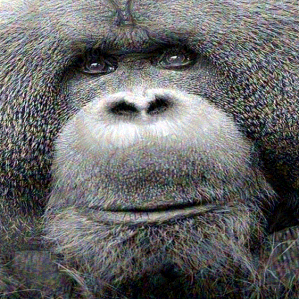} \\[-1.5em]
\rothead{\centering{SI-NI-FGSM}} 
                        &   \includegraphics[width=\hsize,valign=m]{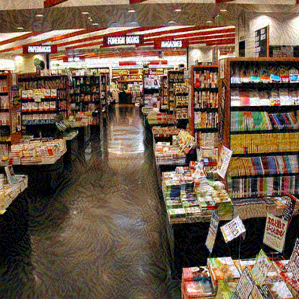}
                        &   \includegraphics[width=\hsize,valign=m]{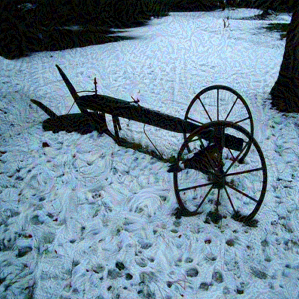}
                        &   \includegraphics[width=\hsize,valign=m]{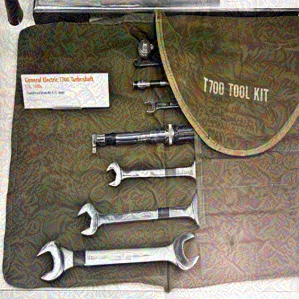}
                        &   \includegraphics[width=\hsize,valign=m]{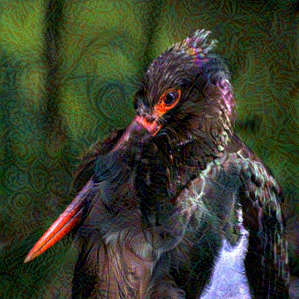}
                        &   \includegraphics[width=\hsize,valign=m]{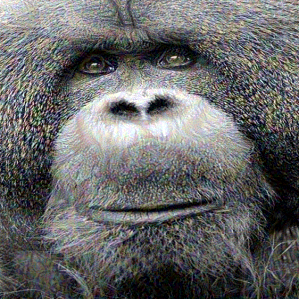}\\[-1.5em]
\rothead{\centering{DA-MI-FGSM}} 
                        &   \includegraphics[width=\hsize,valign=m]{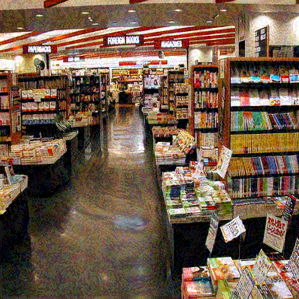}
                        &   \includegraphics[width=\hsize,valign=m]{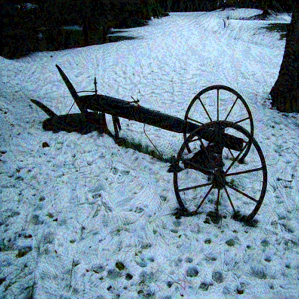}
                        &   \includegraphics[width=\hsize,valign=m]{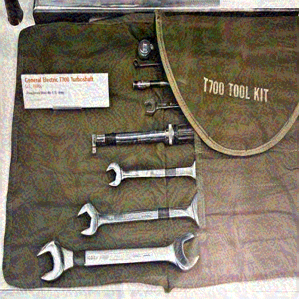}
                        &   \includegraphics[width=\hsize,valign=m]{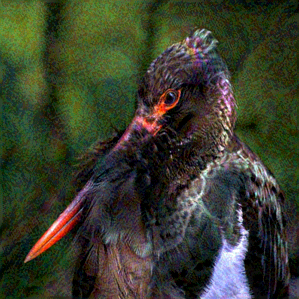}
                        &   \includegraphics[width=\hsize,valign=m]{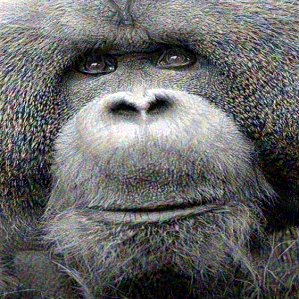}\\[-1.5em]

\end{tabularx}
    \caption{Visualization of randomly selected clean images and their corresponding adversarial examples. All examples are generated by TIM, DIM, SI-NI-FGSM and DA-MI-FGSM attacks respectively.}
\label{Fig:vis}
\end{figure}

\textbf{Visibility.} We visualize 5 randomly selected pairs of adversarial examples generated by TIM, DIM, SI-NI-FGSM and DA-MI-FGSM attacks respectively and their corresponding clean images in Fig.~\ref{Fig:vis}. We can see that the adversarial examples generated by our method are similar to those generated by other methods in visibility, and all these adversarial examples are hard to be distinguished from their corresponding clean images by humans. 

\begin{table*}[htb]
\caption{The attack success rates ($\%$) against Inc-V3, Inc-V4, IncRes-V2, Res-101, Inc-V3$_{ens3}$, Inc-V3$_{ens4}$ and IncRes-V2$_{ens}$ models. The adversarial examples are generated based on Inc-V3 model by normal adversarial attacks including FGSM, I-FGSM, PGD, C\&W and transferable adversarial attacks including DIM, TIM, SI-NI-FGSM, DA-FGSM, DA-I-FGSM and DA-MI-FGSM attacks. $*$ denotes the white-box model being attacked.}
\centering  
\begin{adjustbox}{width=0.9\textwidth}
\begin{tabular}{llccccccc} \\
\thickhline
&Attack & Inc-V$3^*$ &Inc-V4 & IncRes-V2 & Res-101 &Inc-V$3_{ens3}$ &Inc-V$3_{ens4}$ &IncRes-V$2_{ens}$\\
\hline 
\multirow{5}{*}{Normal}&FGSM&67.1 &26.7 &25 &24.4 &10.5 &10 &4.5\\
                        &   I-FGSM &99.9	&20.7	&18.5	&15.3	&3.6	&5.8	&2.9\\
                        &PGD       &99.5	&17.3	&15.1	&13.1	&6.1	&5.6	&3.1\\
                        &C\&W      &100	&18.4	&16.2	&14.3	&3.8	&4.7	&2.7  \\
\hline
\multirow{6}{*}{Transferable}
                       &MI-FGSM &100.0	&40.0	&38.2	&32.3	&12.5	&12.8	&6.8\\
                       &DIM &98.7	&67.7	&62.9	&54	&20.5	&18.4	&9.7 \\
                       &TIM &100	&47.8	&42.8	&39.5	&24	&21.4	&12.9 \\
                       &SI-NI-FGSM  &100	&76	&73.3	&67.6	&31.6	&30	&17.4\\

                       &\textbf{DA-FGSM(Ours)}&87.6&47&43.6&42&18.3&17.4&9.5\\
                       &\textbf{DA-I-FGSM(Ours)}&99.8&44&39.2&34.3&23.7&22.4&12.4\\
                       &\textbf{DA-MI-FGSM(Ours)}&99.8&\textbf{80.6}&\textbf{78.5}&\textbf{72.2}&\textbf{40.6}&\textbf{40.4}&\textbf{26.5}\\
\thickhline
\end{tabular}
\end{adjustbox}
\label{tab:1}
\end{table*}

\begin{table*}[htb]
\caption{Comparison of TIM, SI-NI-TIM and the DA-TIM extension. The attack success rates ($\%$) are shown in the table. Adversarial examples are generated based on Inc-V3, Inc-V4, IncRes-V2 and Res-101 respectively. $*$ denotes the attack success rates under white-box attacks. }
\centering  
\begin{adjustbox}{width=0.9\textwidth}
\begin{tabular}{llccccccc} \\
\thickhline
Model&Attack & Inc-V3 &Inc-V4 & IncRes-V2 & Res-101 &Inc-V$3_{ens3}$ &Inc-V$3_{ens4}$ &IncRes-V$2_{ens}$\\
\hline 
\multirow{3}{*}{Inc-V3}&TIM &100$^*$	&47.8	&42.8	&39.5	&24	&21.4	&12.9\\
                       &SI-NI-TIM&100$^*$	&77.2	&75.8	&66.5	&51.8	&45.9	&33.5\\
                       &\textbf{DA-TIM(Ours)}&99.8$^*$&\textbf{80.9}&\textbf{77.9}&\textbf{71.8}&\textbf{66.9}&\textbf{65.2}&\textbf{51.2}\\
                       
\hline
\multirow{3}{*}{Inc-V4}&TIM&58.5	&99.6$^*$	&47.5	&43.2	&25.7	&23.3	&17.3
\\
                       &SI-NI-TIM&83.5	&100$^*$	&76.6	&68.9	&57.8	&54.3	&42.9
\\
                       &\textbf{DA-TIM(Ours)}&\textbf{84.2}&98.4$^*$&\textbf{77.7}&\textbf{69.3}&\textbf{66.8}&\textbf{65.9}&\textbf{56.4}\\
                       
\hline
\multirow{3}{*}{IncRes-V2}&TIM&62	&56.2	&97.5$^*$	&51.3	&32.8	&27.9	&21.9\\
                       &SI-NI-TIM&\textbf{86.4}	&\textbf{83.2}	&99.5$^*$	&\textbf{77.2}	&66.1	&60.2	&57.1\\
                       &\textbf{DA-TIM(Ours)}&80&78.5&94$^*$&74&\textbf{69.5}&\textbf{66.4}&\textbf{66}\\
                       
\hline
\multirow{3}{*}{Res-101}&TIM&59	&53.6	&51.8	&99.3$^*$	&36.8	&32.2	&23.5
\\
                       &SI-NI-TIM &78.3	&74.1	&73	&99.8$^*$	&58.9	&53.9	&43.1
\\
                       &\textbf{DA-TIM(Ours)}&\textbf{78.6}&\textbf{74.7}&\textbf{76}&99.2$^*$&\textbf{72.1}&\textbf{69.7}&\textbf{62.7}\\
                       
\thickhline
\end{tabular}
\end{adjustbox}
\label{tab:2}
\end{table*}

\begin{table*}[htb]
\caption{Comparison of DIM, SI-NI-DIM and the DA-DIM extension. The numbers in table denote the attack success rates ($\%$). Adversarial examples are generated based on Inc-V3, Inc-V4, IncRes-V2 and Res-101 respectively using DIM, SI-NI-DIM and DA-DIM methods. $*$ denotes the attack success rates under white-box attacks.}
\centering  
\begin{adjustbox}{width=0.9\textwidth}
\begin{tabular}{llccccccc} \\
\thickhline
Model&Attack & Inc-V3 &Inc-V4 & IncRes-V2 & Res-101 &Inc-V$3_{ens3}$ &Inc-V$3_{ens4}$ &IncRes-V$2_{ens}$\\
\hline 
\multirow{3}{*}{Inc-V3}&DIM&98.7$^*$	&67.7	&62.9	&54	&20.5	&18.4	&9.7
\\
                       &SI-NI-DIM&99.6$^*$	&84.7	&81.7	&75.4	&36.9	&34.6	&20.2
\\
                       &\textbf{DA-DIM(Ours)}&99.5$^*$&\textbf{89}&\textbf{87.3}&\textbf{81.2}&\textbf{57.1}&\textbf{56.6}&\textbf{38.8}\\
\hline
\multirow{3}{*}{Inc-V4}&DIM&70.7	&98.0$^*$	&63.2	&55.9	&21.9	&22.3	&11.9
\\
                       &SI-NI-DIM&89.7	&99.3$^*$	&84.5	&78.5	&47.6	&45	&28.9
\\
                       &\textbf{DA-DIM(Ours)}&\textbf{90.8}&98.1$^*$&\textbf{87.1}&\textbf{80.9}&\textbf{62.1}&\textbf{62.9}&\textbf{49.7}\\
                       
\hline
\multirow{3}{*}{IncRes-V2}&DIM&69.1	&63.9	&93.6$^*$	&47.4	&29.4	&24	&17.3
\\
                       &SI-NI-DIM&\textbf{89.7}	&\textbf{86.4}	&99.1$^*$	&\textbf{81.2}	&55	&48.2	&38.1
\\
                       &\textbf{DA-DIM(Ours)}&86.1&85.8&95$^*$&80.2&\textbf{64.6}&\textbf{59.7}&\textbf{57.1}\\
                       
\hline
\multirow{3}{*}{Res-101}&DIM&75.9	&70	&71	&98.3$^*$	&36	&32.4	&19.3
\\
                       &SI-NI-DIM&88.7	&84.2	&84.4	&99.3$^*$	&53.4	&48	&33.2
\\
                       &\textbf{DA-DIM(Ours)}&\textbf{90.9}&\textbf{87.7}&\textbf{89.4}&99.2$^*$&\textbf{75.3}&\textbf{72.6}&\textbf{62.9}\\
                       
\thickhline
\end{tabular}
\end{adjustbox}
\label{tab:3}
\end{table*}
\begin{table*}[htb]
\caption{Comparison of TI-DIM, SI-NI-TI-DIM and the DA-TI-DIM extension. The numbers in table denote the attack success rates ($\%$). Adversarial examples are generated based on Inc-V3, Inc-V4, IncRes-V2 and Res-101 respectively using TI-DIM, SI-NI-TI-DIM and DA-TI-DIM methods. $*$ denotes the attack success rates under white-box attacks. }
\centering  
\begin{adjustbox}{width=0.9\textwidth}
\begin{tabular}{llccccccc} \\
\thickhline
Model&Attack & Inc-V3 &Inc-V4 & IncRes-V2 & Res-101 &Inc-V$3_{ens3}$ &Inc-V$3_{ens4}$ &IncRes-V$2_{ens}$\\
\hline 
\multirow{3}{*}{Inc-V3}&TI-DIM&98.5$^*$	&66.1	&63	&56.1	&38.6	&34.9	&22.5
\\
                       &SI-NI-TI-DIM&99.6$^*$	&85.5	&80.9	&75.7	&61.5	&56.9	&40.7
\\
                       &\textbf{DA-TI-DIM(Ours)}&99.6$^*$&\textbf{88.3}&\textbf{85.1}&\textbf{80.3}&\textbf{77.4}&\textbf{76.8}&\textbf{62.9}\\
                       
\hline
\multirow{3}{*}{Inc-V4}&TI-DIM&72.5	&97.8$^*$	&63.4	&54.5	&38.1	&35.2	&25.3
\\
                       &SI-NI-TI-DIM&88.1	&99.3$^*$	&83.7	&77	&65	&63.1	&49.4
\\
                       &\textbf{DA-TI-DIM(Ours)}&\textbf{88.8}&97.8$^*$&\textbf{83.9}&\textbf{78.3}&\textbf{75.7}&\textbf{75.7}&\textbf{68.1}\\
                       
\hline
\multirow{3}{*}{IncRes-V2}&TI-DIM&73.2	&67.5	&92.4$^*$	&61.3	&46.4	&40.2	&35.8
\\
                       &SI-NI-TI-DIM&\textbf{89.6}	&\textbf{87}	&99.1$^*$	&\textbf{83.9}	&74	&67.9	&63.7
\\
                       &\textbf{NS-TI-DIM(Ours)}&84.2&83.5&94.5$^*$&78.3&\textbf{76.1}&\textbf{73.1}&\textbf{72.8}\\
                       
\hline
\multirow{3}{*}{Res-101}&TI-DIM&74.9	&69.8	&70.5	&98.7$^*$	&52.6	&49.1	&37.8
\\
                       &SI-NI-TI-DIM&86.4	&82.6	&84.6	&99$^*$	&72.6	&66.8	&56.4
\\
                       &\textbf{DA-TI-DIM(Ours)}&\textbf{88.1}&\textbf{83.8}&\textbf{86.2}&99.3$^*$&\textbf{82.6}&\textbf{82.2}&\textbf{76.2}\\
                       
\thickhline
\end{tabular}
\end{adjustbox}
\label{tab:4}
\end{table*}

\subsection{Ensemble-based Attacks}
We also evaluate the performance of our method under ensemble-based attacks. Liu et al.~\cite{liu2016delving} have shown that attacking multiple models simultaneously can generate more transferable adversarial examples. It is because if an adversarial example can attack multiple models successfully, it can more likely attack yet another model successfully. 

We follow the ensemble-based attack strategy proposed in~\cite{dong2018boosting}, which fuses the logit activations of multiple models to generate adversarial examples. In this experiment, we generate adversarial examples by attacking Inc-V3, Inc-V4, IncRes-V2 and Res-101 models simultaneously with equal ensemble weights. 
In Table~\ref{tab:5}, we show the attack success rates for DA-DIM, DA-TIM, DA-TI-DIM attacks and baselines. It shows that our method outperforms these baselines across all experiments. The highest attack success rate is achieved by our DA-TI-DIM attack and the average attack success rates against the three robust models reach 94.6\%. 

\begin{table*}[htb]
\caption{The attack success rates ($\%$) against Inc-V3, Inc-V4, IncRes-V2, Res-101, Inc-V3$_{ens3}$, Inc-V3$_{ens4}$ and IncRes-V2$_{ens}$ models. Adversarial examples are generated based on the ensemble of Inc-V3, Inc-V4, IncRes-V2 and Res-101 models using DIM, SI-NI-DIM, TIM, SI-NI-TIM, TI-DIM, SI-NI-TI-DIM, DA-DIM, DA-TIM and DA-TI-DIM attacks respectively. $Average$ column denotes the averaged attack success rates against the three robust models. $*$ denotes the white-box model being attacked.}
\centering  
\begin{adjustbox}{width=0.9\textwidth}
\begin{tabular}{lcccccccc} \\
\thickhline
Attack & Inc-V$3^*$ &Inc-V$4^*$ & IncRes-V$2^*$ & Res-10$1^*$ &Inc-V$3_{ens3}$ &Inc-V$3_{ens4}$ &IncRes-V$2_{ens}$ & Average\\
\hline 
DIM&99.7	&99.2	&98.9	&98.9	&66.4	&60.9	&41.6 &56.3\\
SI-NI-DIM &100	&100	&100	&99.9	&88.2	&85.1	&69.7 &81\\
\textbf{DA-DIM(Ours)}       &99.9	&99.8	&99.7	&99.8	&\textbf{91}	&\textbf{90.1}	&\textbf{85.5} &\textbf{88.9}\\
\hline
TIM      &99.9	&99.3	&99.3	&99.8	&71.6	&67	&53.2 &63.9\\
SI-NI-TIM      &100	&100	&100	&100	&93.2	&90.1	&84.5 &89.2\\
\textbf{DA-TIM(Ours)}      &99.8	&99.8	&99.2	&99.6	&\textbf{93.4}	&\textbf{92.1}	&\textbf{89.3} &\textbf{91.6}\\
\hline
TI-DIM      &99.6	&98.8	&98.8	&98.9	&85.2	&80.2	&73.3 &79.5
  \\
SI-NI-TI-DIM     &99.9	&99.9	&99.9	&99.9	&96	&94.3	&90.3 &93.5
  \\
\textbf{DA-TI-DIM(Ours)}      &99.8	&99.8	&99.6	&99.6	&\textbf{96.2}	&\textbf{94.7}	&\textbf{93} &\textbf{94.6}\\
\thickhline
\end{tabular}
\end{adjustbox}
\label{tab:5}
\end{table*}
\subsection{Attacking Other Defense Models}
We also study the performance of our method on defense models. We test it against HGD~\cite{liao2018defense}, R\&P~\cite{xie2017mitigating}, NIPS-r3, FD~\cite{liu2019feature} and ComDefend~\cite{jia2019comdefend} defense methods. HGD, R\&P and NIPS-r3 were the top 3 defense methods in NIPS 2017 defense competition. FD and ComDefend are recently published defense methods for purifying adversarial perturbations. TI-DIM~\cite{dong2019evading} and SI-NI-TI-DIM attacks~\cite{Lin2020Nesterov} are presented as baselines. Adversarial examples are generated based on the ensemble of Inc-V3, Inc-V4, IncRes-V2 and Res-101 models. The attack success rates against FD and ComDefend defense are based on IncRes-V2$_{ens}$ model. 

As shown in Table~\ref{tab:6}, our model achieves state-of-the-art results and reaches 94.8\% for averaged attack success rates, which indicates current defense methods are not safe to transferable adversarial attacks.
\begin{table}[htb]
\caption{The attack success rates against the five advanced defense models.}

\centering  
\begin{tabular}{lcccccc} \\
\thickhline
Attack & HGD &R\&P & NIPS-r3 & FD &ComDefend &Average\\
\hline 
TI-DIM&84.8	&75.3	&80.7	&84.2	&79.6		&80.9\\
SI-NI-TI-DIM &96.1	&91.3	&94.4	&93.7	&91.9	&93.5\\
\textbf{DA-TI-DIM(Ours)}    &\textbf{96.1}	&\textbf{93.6}	&\textbf{94.8}	&\textbf{94.4}	&\textbf{94.3}	&\textbf{94.8}\\
\thickhline
\end{tabular}
\label{tab:6}
\end{table}

\subsection{Similarity of adversarial  perturbations}
\begin{figure*}[htb]
        \centering
        \subfloat{
            \includegraphics[width=0.9\textwidth]{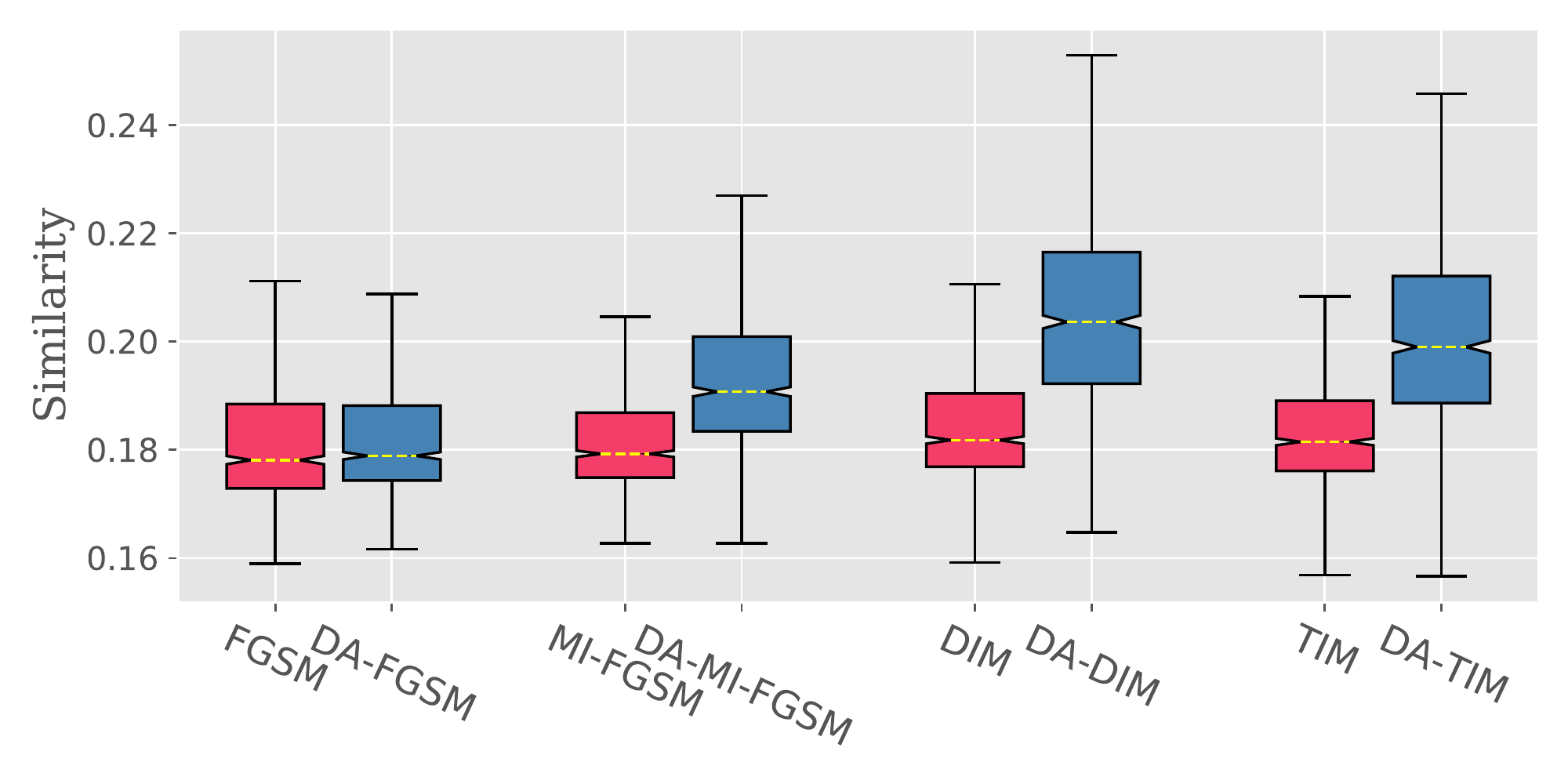}
        }
    \caption{The cosine similarity of adversarial  perturbations generated from Inc-V3, Inc-V4, IncRes-V2 and Res-101 models.}
    \label{fig:Similarity}
\end{figure*}

To further understand the proposed Direction-Aggregated attack, we plot the cosine similarity of adversarial  perturbations generated from multiple white-box models, i.e. Inc-V3, Inc-V4, IncRes-V2 and Res-101 models. The results are showed in Fig~\ref{fig:Similarity}.

In Fig~\ref{fig:Similarity}, the cosine similarity of adversarial  perturbations generated by the propose Direction-Aggregated attack is generally higher than other baseline attacks. It is in line with our expectation since the aggregated direction could reduce the oscillation of each update direction in generating adversarial perturbations. Besides, we notice that the cosine similarity of adversarial  perturbations on DA-FGSM is not significant higher than FGSM. We conjecture that it is due to the adversarial  perturbations generated by FGSM ``underfit'' the white-box model, which limits the similarity of adversarial  perturbations.       

\subsection{Parameter Analysis}
In this section, we conduct a series of experiments to study the impact of different hyper-parameters on the transferability of adversarial examples.

\begin{figure*}[htb]
        \centering
        \subfloat[Inc-V3]{
            \includegraphics[width=0.46\textwidth]{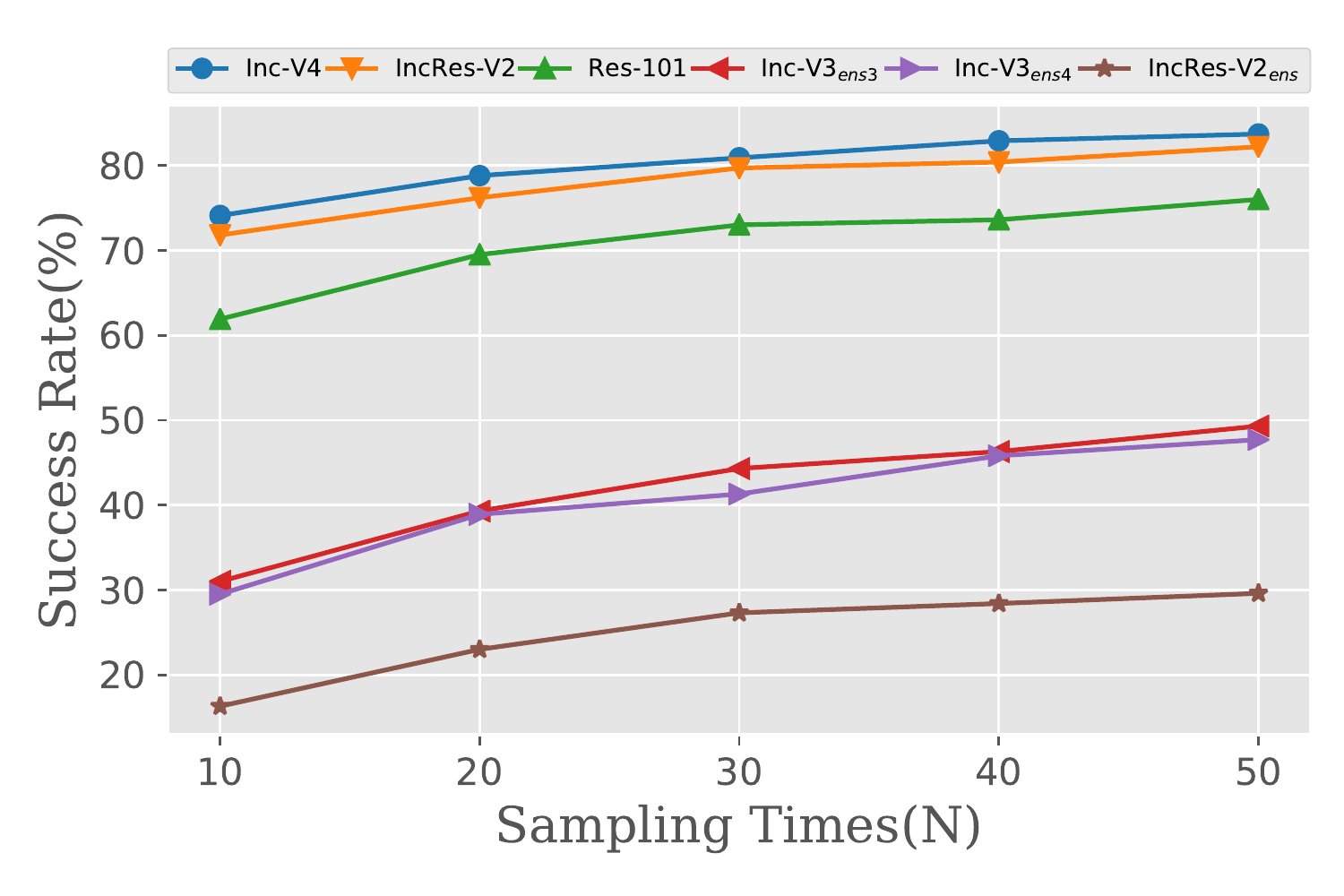}
            \label{fig:2a}
            
        }
    \subfloat[Inc-v4]{
        \includegraphics[width=0.46\textwidth]{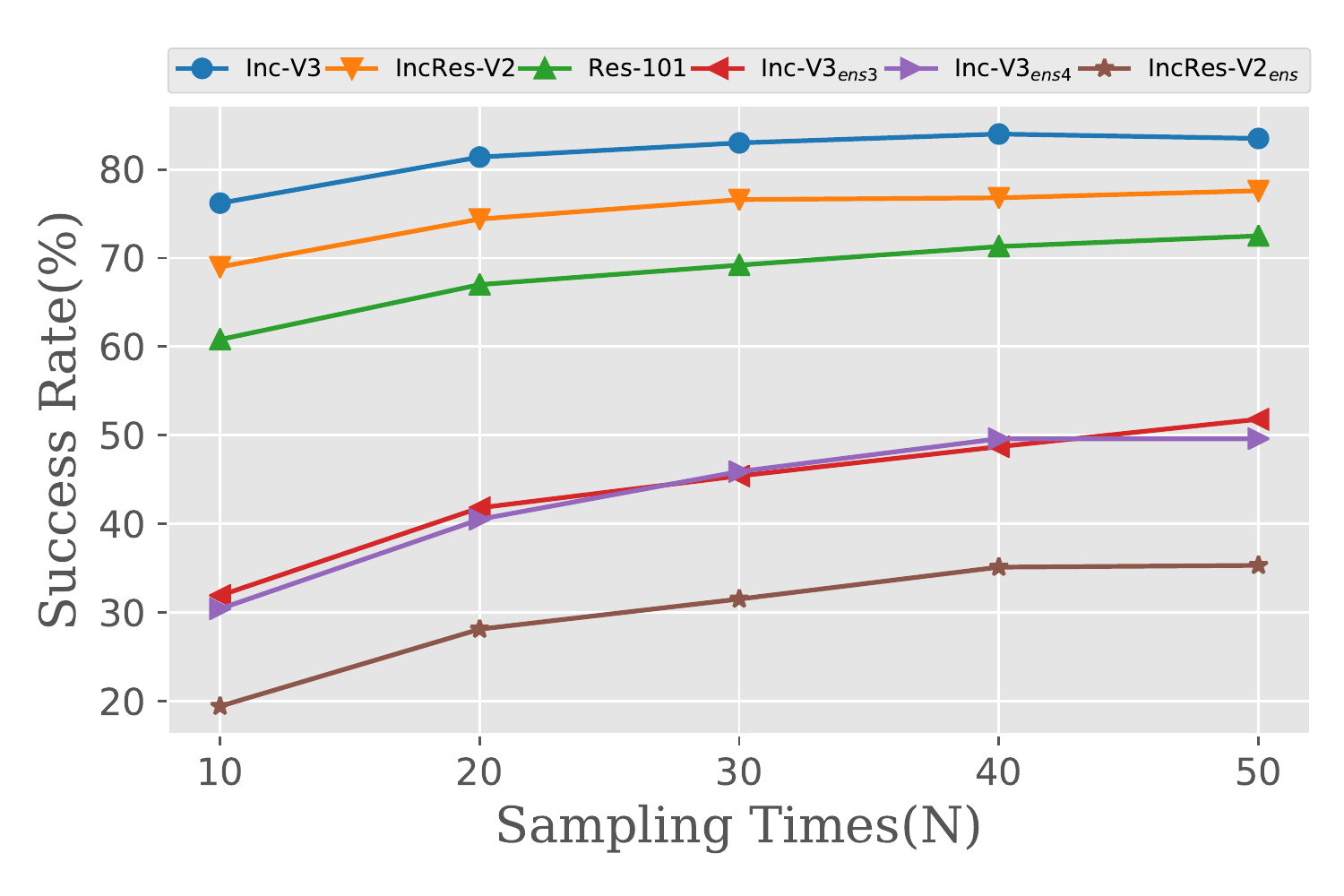}
        \label{fig:2b}
    }\\
        \subfloat[IncRes-V2]{
        \includegraphics[width=0.46\textwidth]{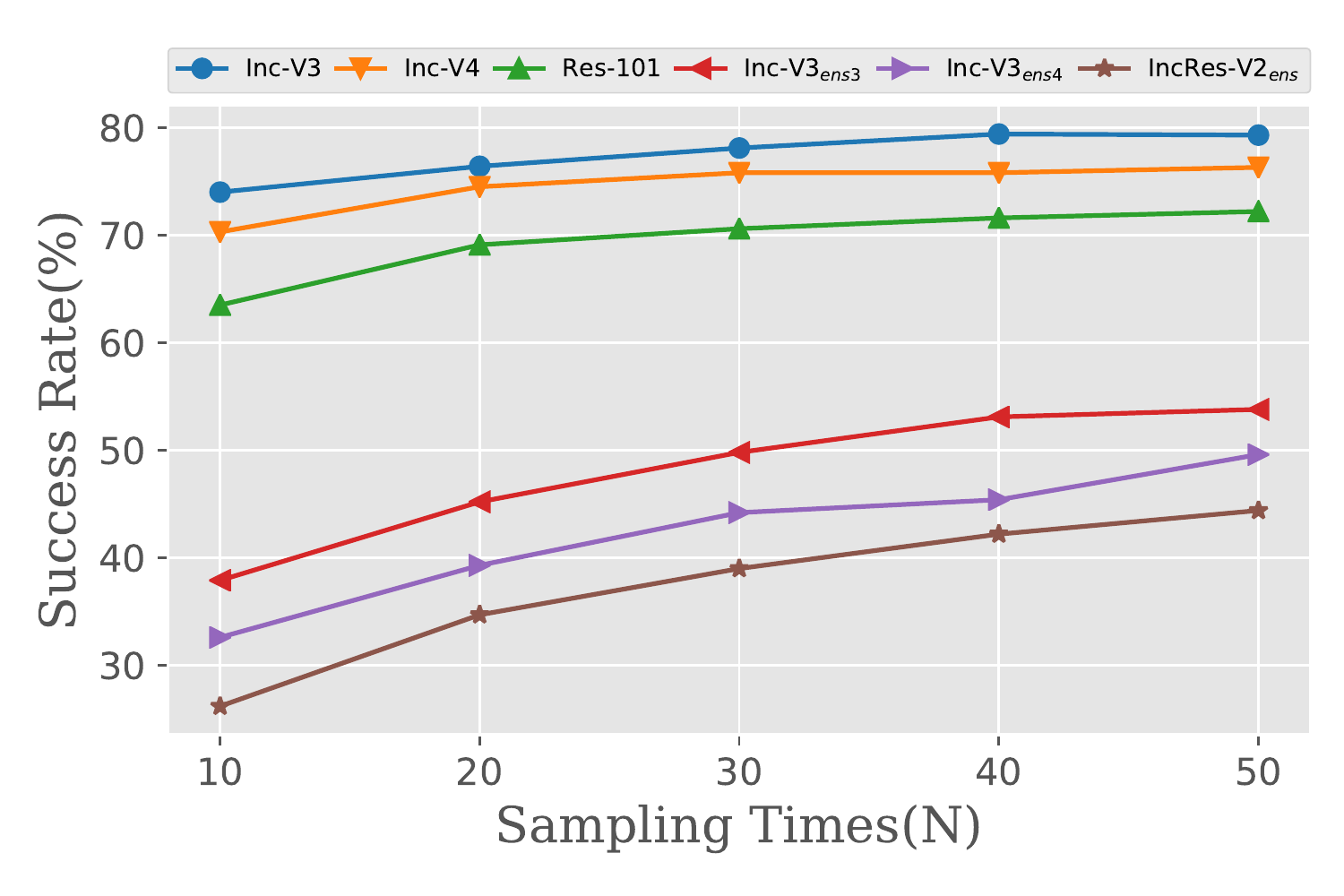}
        \label{fig:2c}
        
    }
    \subfloat[Res-101]{
        \includegraphics[width=0.46\textwidth]{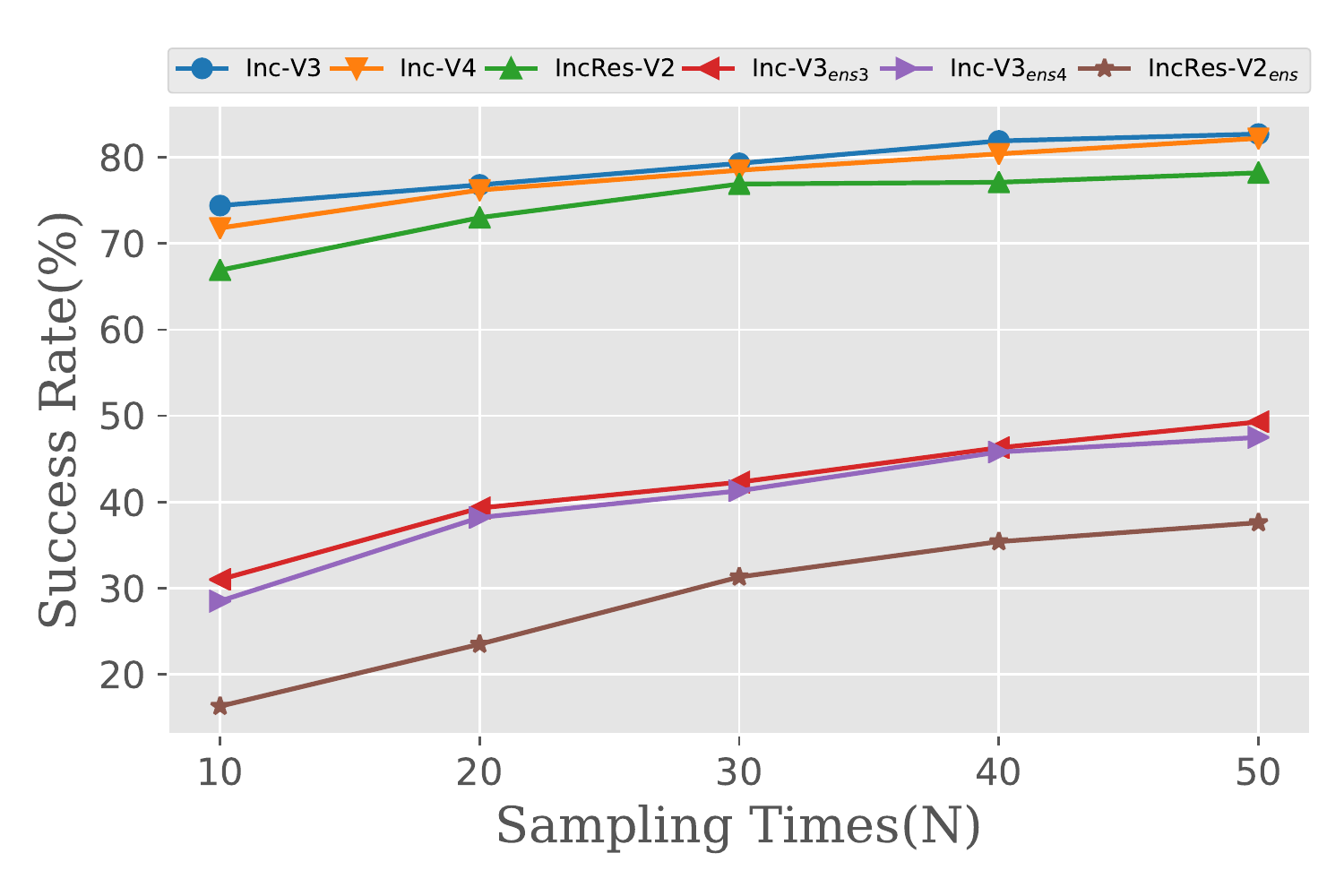}
        \label{fig:2d}
    }
    \caption{The attack success rates ($\%$) of black-box attack against Inc-V3, Inc-V4, IncRes-V2, Res-101, Inc-V$3_{ens3}$, Inc-V$3_{ens4}$ and IncRes-V$2_{ens}$ models when varying sampling times $N$ ranging from 10 to 50. The adversarial examples are generated based on Inc-V3 (Fig.~\ref{fig:2a}), Inc-V4 (Fig.~\ref{fig:2b}), IncRes-V2 (Fig.~\ref{fig:2c}) and Res-101 (Fig.~\ref{fig:2d}) models respectively by DA-MI-FGSM attack.}
    \label{fig:2}
\end{figure*}

\textbf{Sampling Times $N$.} We explore the influence of sampling times $N$ upon the transferability of adversarial examples. Fig.~\ref{fig:2} shows the attack success rates (\%) against Inc-V3, Inc-V4, IncRes-V2, Res-101, Inc-V3$_{ens3}$, Inc-V3$_{ens4}$ and IncRes-V2$_{ens}$ models under black-box settings. The generation of adversarial examples is based on Inc-V3, Inc-V4, IncRes-V2 and Res-101 models respectively with standard deviation $\sigma$ setting as $0.05$. 

From Fig.~\ref{fig:2}, we can see that the attack success rates are growing with the increase of sampling times. In detail, the curve is growing fast when sampling times $N$ is less than 30 and the trend of growth tends to be flattening when sampling times $N$ is greater than 30. Besides, the growing trends of Fig.~\ref{fig:2a}, Fig.~\ref{fig:2b}, Fig.~\ref{fig:2c} and Fig.~\ref{fig:2d} are similar, which indicates that the influence of sampling times $N$ on the transferability is little sensitive to the white-box model. 

\begin{figure*}[htb]
    \centering
    \subfloat[Inc-V3]{
        \includegraphics[width=0.47\textwidth]{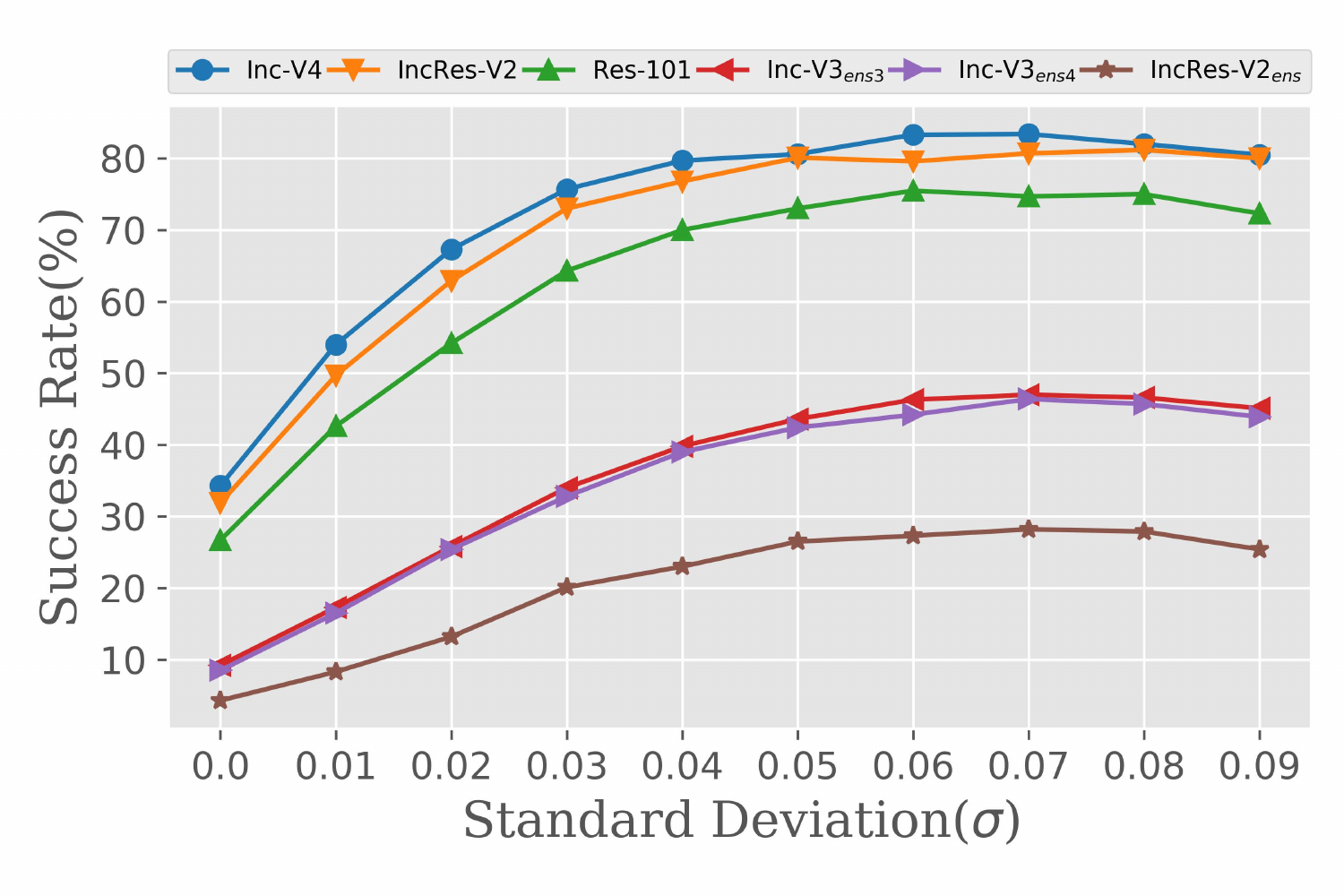}
        \label{fig:3a}
    }
    \subfloat[Inc-V4]{
        \includegraphics[width=0.47\textwidth]{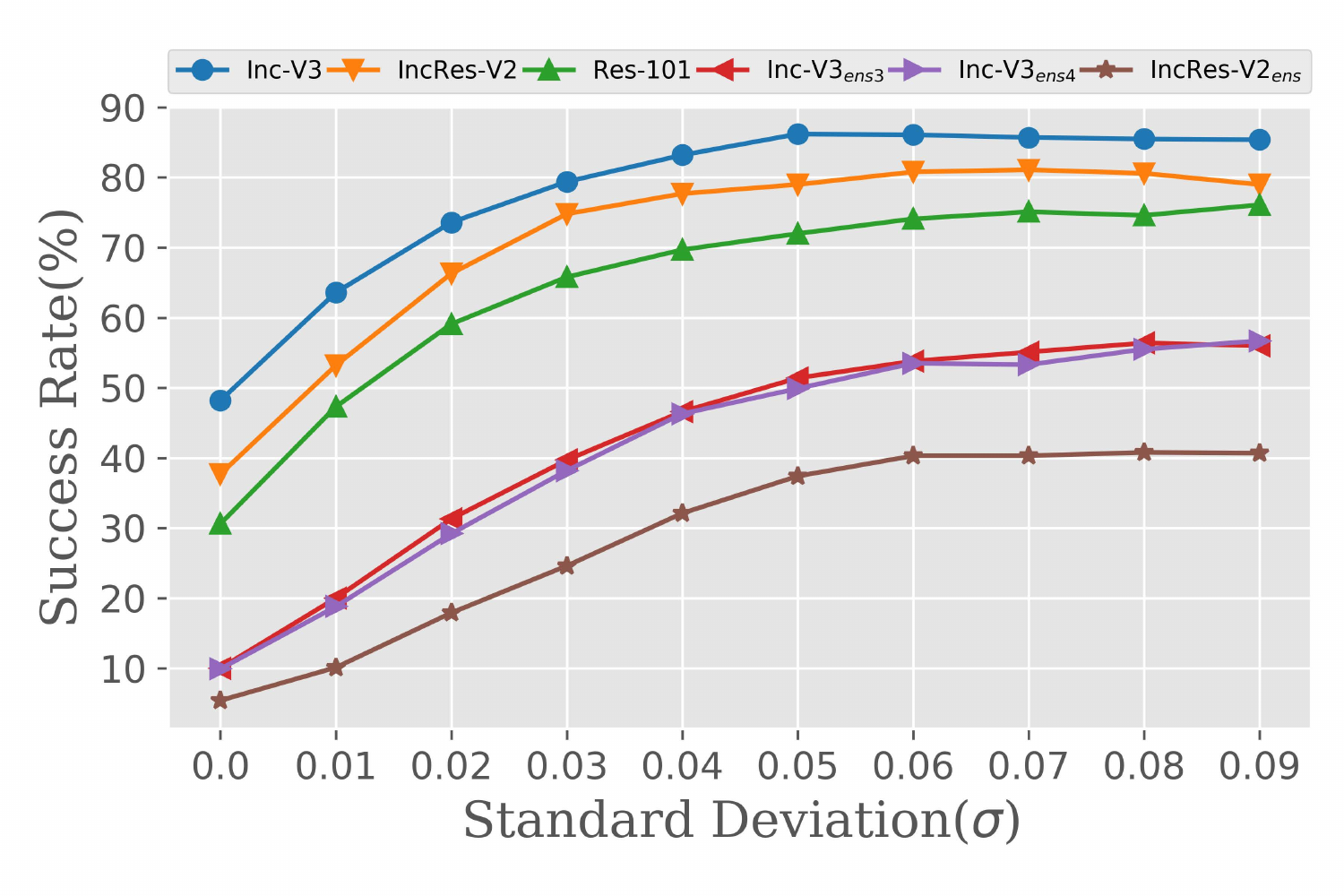}
        \label{fig:3b}
        
    }\\
        \subfloat[IncRes-V2]{
        \includegraphics[width=0.47\textwidth]{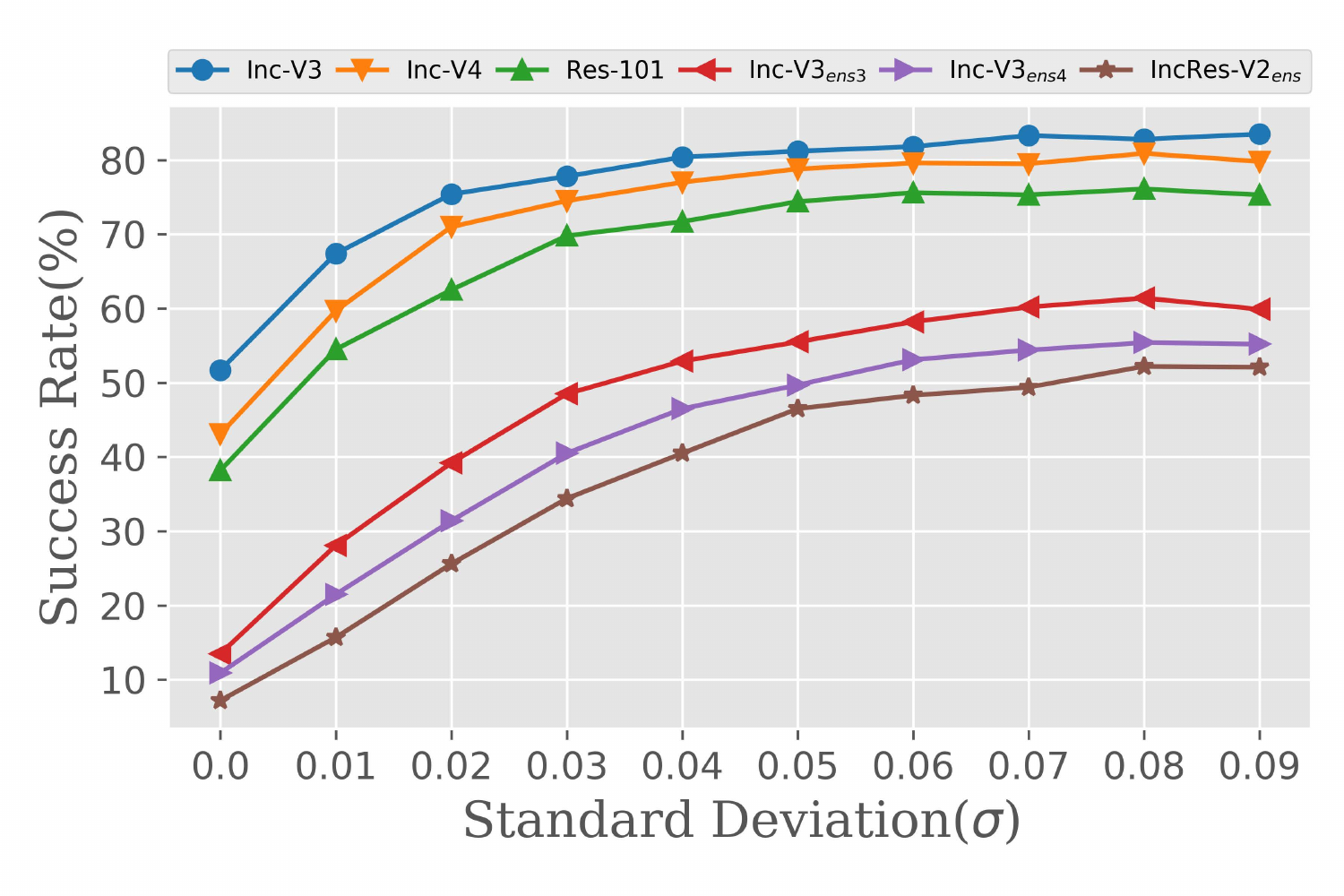}
        \label{fig:3c}
    }
    \subfloat[Res-101]{
        \includegraphics[width=0.47\textwidth]{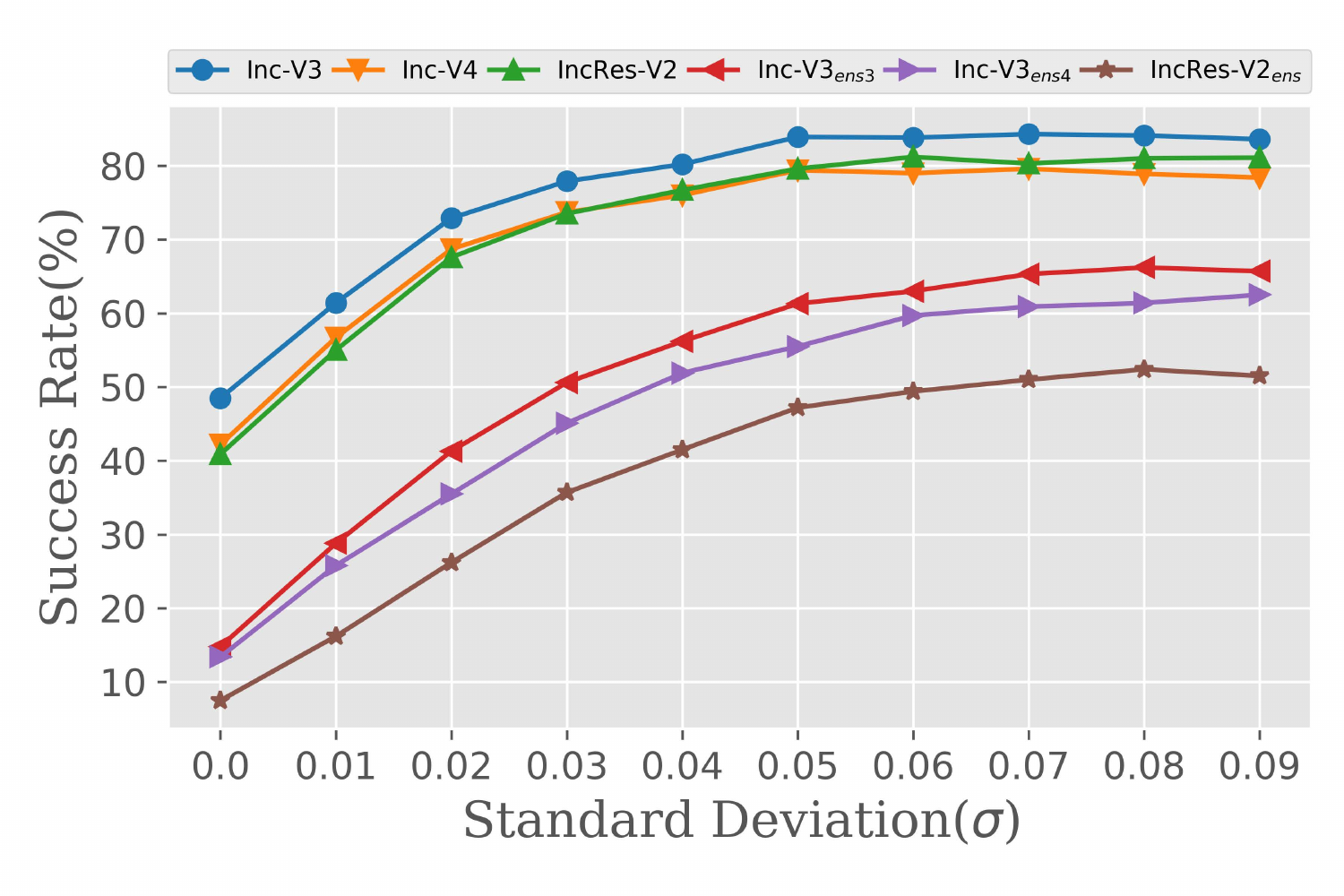}
        \label{fig:3d}
        
    }
    \caption{The attack success rates ($\%$) of black-box attack against Inc-V3, Inc-V4, IncRes-V2, Res-101, Inc-V$3_{ens3}$, Inc-V$3_{ens4}$ and IncRes-V$2_{ens}$ models when varying $\sigma$ from 0 to 0.09. The adversarial examples are generated based on Inc-V3 (Fig.~\ref{fig:3a}), Inc-V4 (Fig.~\ref{fig:3b}), IncRes-V2 (Fig.~\ref{fig:3c}) and Res-101 (Fig.~\ref{fig:3d}) models respectively using DA-MI-FGSM attack.}
    \label{fig:3}
\end{figure*}

\textbf{$\sigma$ in Gaussian Distribution.}
Standard deviation $\sigma$ controls the shape of Gaussian distribution and plays an important role in Gaussian noise generation. We study the influence of $\sigma$ upon the transferability of adversarial examples. Fig.~\ref{fig:3} shows the attack success rates against Inc-V3, Inc-V4, IncRes-V2, Res-101, Inc-V$3_{ens3}$, Inc-V$3_{ens4}$ and IncRes-V$2_{ens}$ models under black-box attacks. Adversarial examples in this experiment are crafted based on Inc-V3, Inc-V4, IncRes-V2 and Res-101 models respectively with sampling times $N=30$.

From Fig.~\ref{fig:3}, we can see that the attack success rates have a surge increasing at first, then the growing trends tend to be flattening. The surge increasing of the attack success rates indicates that the parameter $\sigma$ plays an important role in our method. Besides, the similar trends among Fig.~\ref{fig:3a}, Fig.~\ref{fig:3b}, Fig.~\ref{fig:3c} and Fig.~\ref{fig:3d} indicate that the influence of $\sigma$ on achieving transferability is insensitive to the white-box model.

It is deserved to note that a very large $\sigma$ is not encouraged for our method for the two reasons: 1) a larger $\sigma$ indicates larger perturbation size will be generated (Fig.~\ref{fig:1}), thus more sampling times are needed to cover the sampling region; 2) noise sampling from a very large $\sigma$ might already be too large to flip the prediction and consequently disturb the attack direction.   

\begin{figure*}[htb]
    \centering
    \subfloat[Inc-V3]{
        \includegraphics[width=0.47\textwidth]{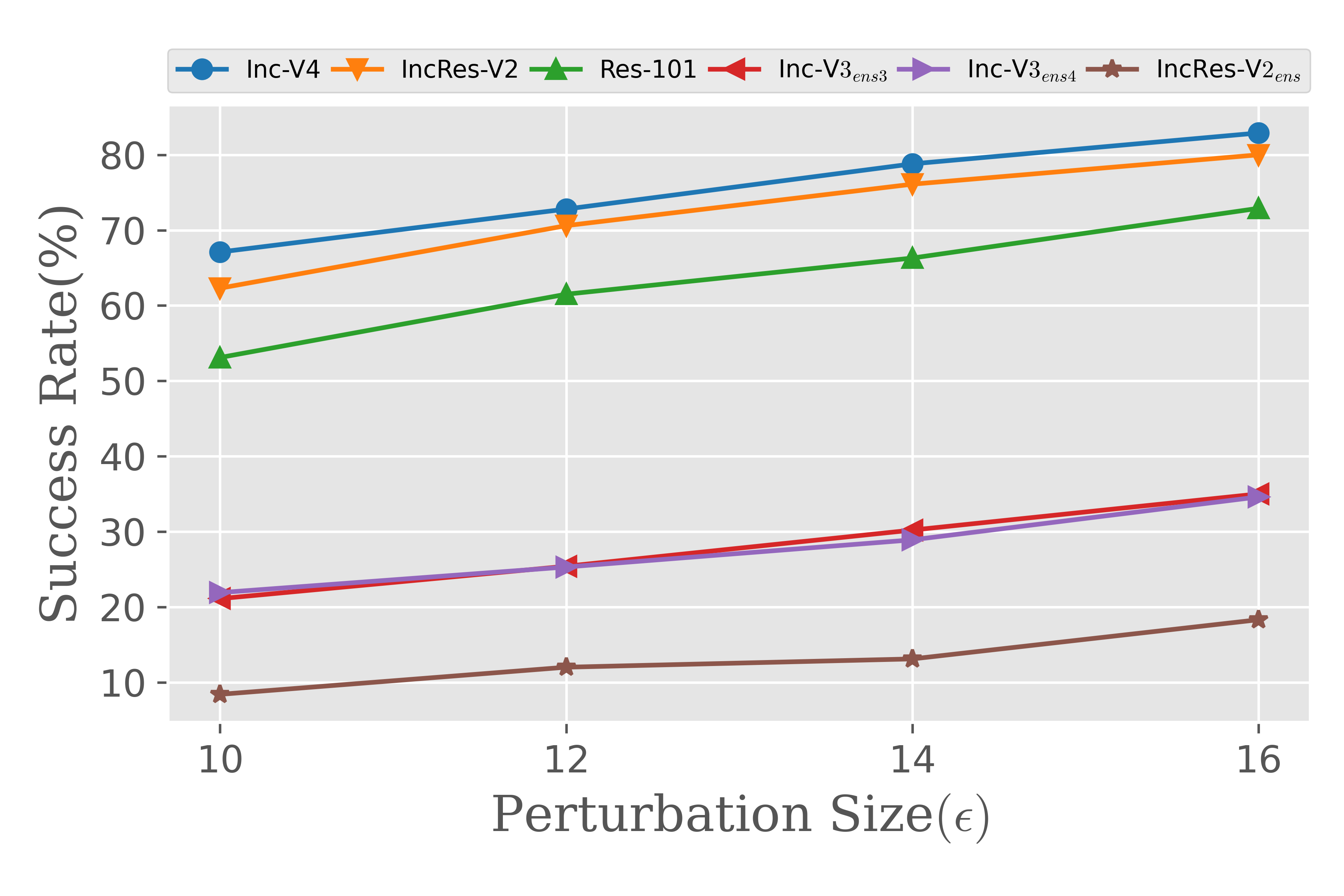}
        \label{fig:4a}
    }
    \subfloat[Inc-V4]{
        \includegraphics[width=0.47\textwidth]{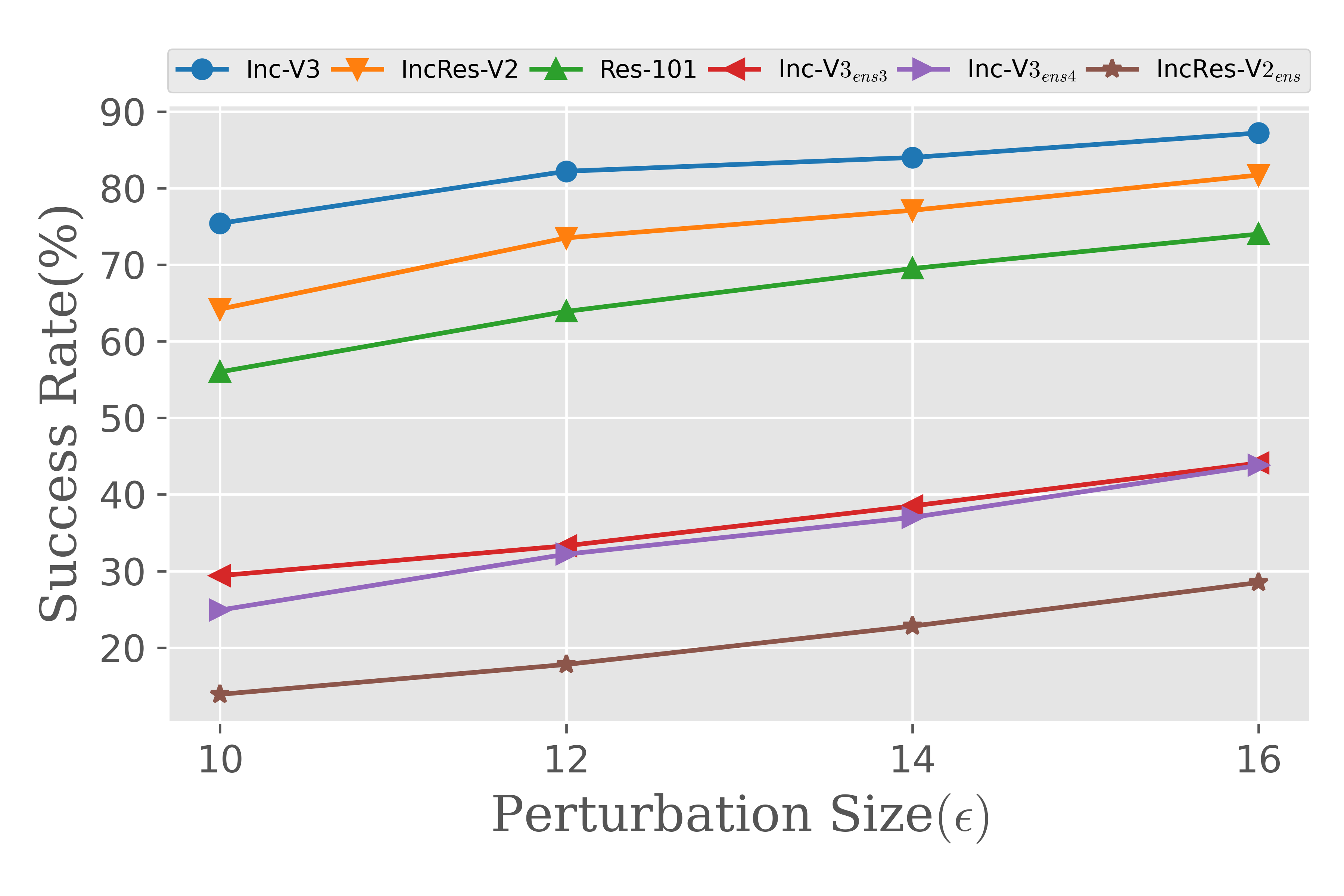}
        \label{fig:4b}
        
    }\\
        \subfloat[IncRes-V2]{
        \includegraphics[width=0.47\textwidth]{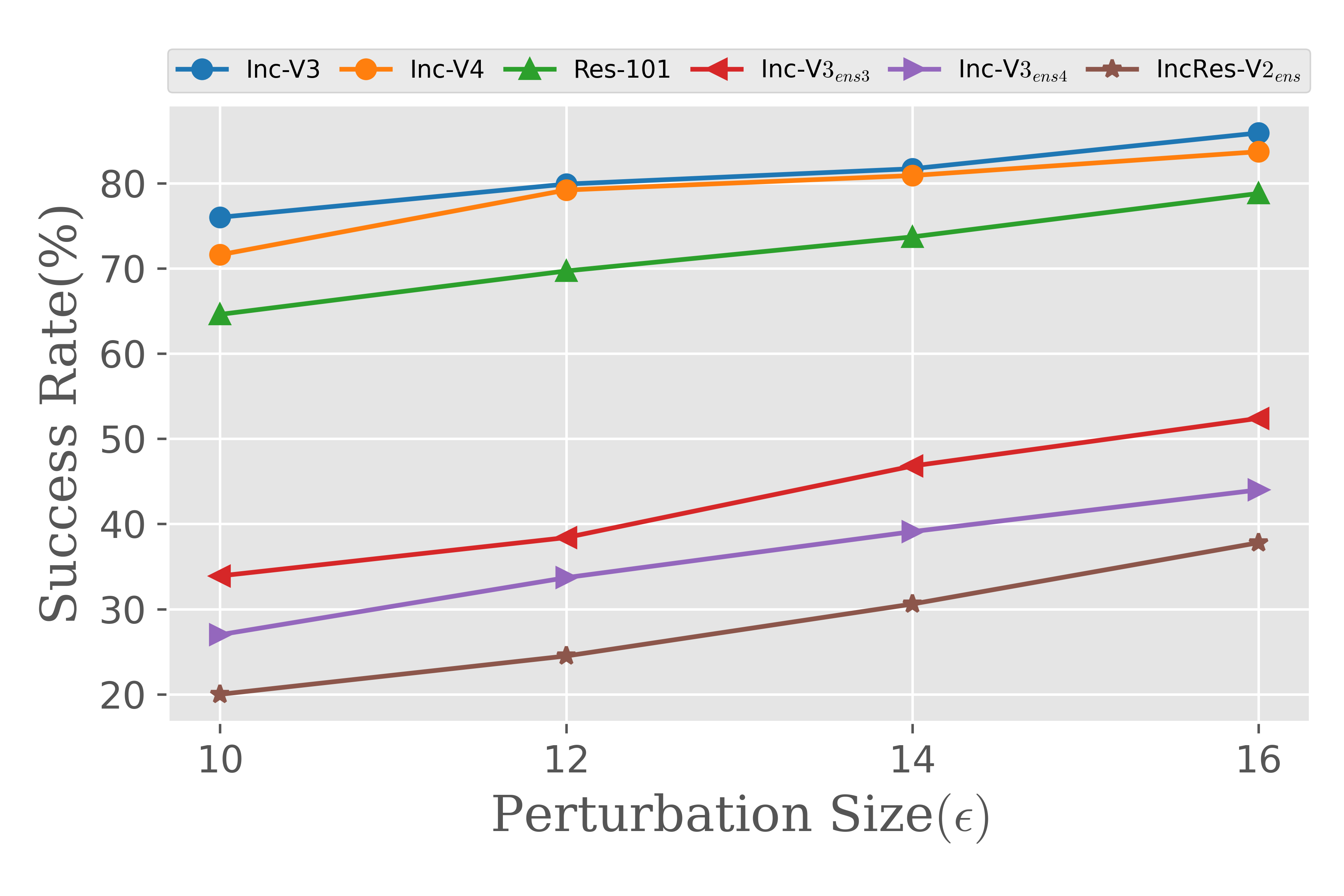}
        \label{fig:4c}
    }
    \subfloat[Res-101]{
        \includegraphics[width=0.47\textwidth]{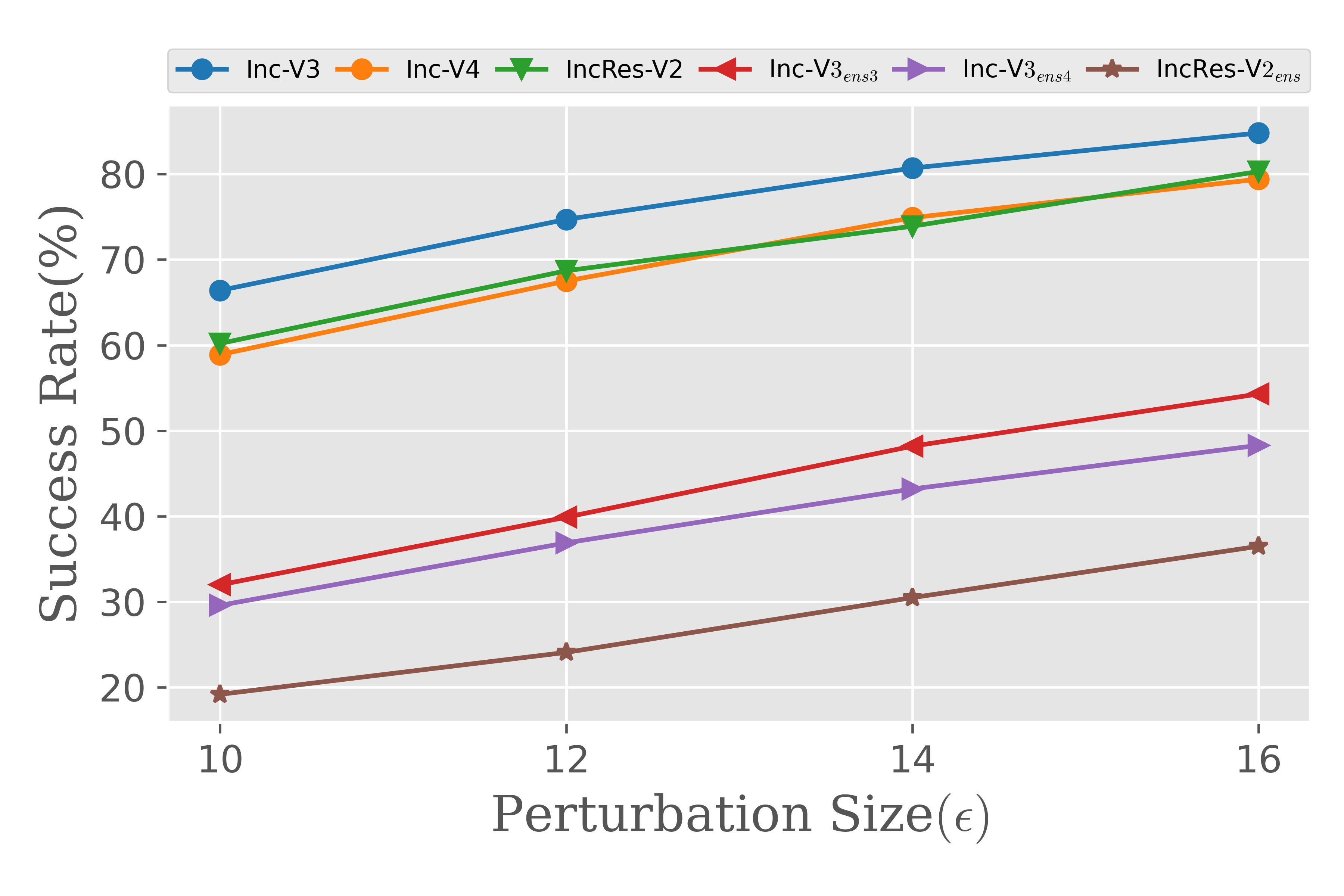}
        \label{fig:4d}
        
    }
    \caption{The attack success rates ($\%$) of black-box attack against Inc-V3, Inc-V4, IncRes-V2, Res-101, Inc-V$3_{ens3}$, Inc-V$3_{ens4}$ and IncRes-V$2_{ens}$ models when varying $\epsilon$ from 10 to 16. The adversarial examples are generated based on Inc-V3 (Fig.~\ref{fig:4a}), Inc-V4 (Fig.~\ref{fig:4b}), IncRes-V2 (Fig.~\ref{fig:4c}) and Res-101 (Fig.~\ref{fig:4d}) models respectively using DA-MI-FGSM attack.}
    \label{fig:4}
\end{figure*}

\textbf{Perturbation Size $\epsilon$.} 
We study the impact of perturbation size $\epsilon$ on the attack success rates. We set sampling times $N$ and standard deviation $\sigma$ to 30 and 0.05 respectively. We fix step size $alpha$ to $\frac{16}{10}$ and iterations $T$ to 16. The attack success rates ($\%$) against Inc-V3, Inc-V4, IncRes-V2, Res-101, Inc-V$3_{ens3}$, Inc-V$3_{ens4}$ and IncRes-V$2_{ens}$ models are achieved under black-box settings. The $\epsilon$ varies from 10 to 16 and the results are showed in Fig.~\ref{fig:4}.

From Fig.~\ref{fig:4}, we observe that the attack success rates increase steadily as perturbation size $\epsilon$ increases on both adversarial trained models and normal trained models. 

\begin{figure*}[htb]
    \centering
    \subfloat[Inc-V3]{
        \includegraphics[width=0.47\textwidth]{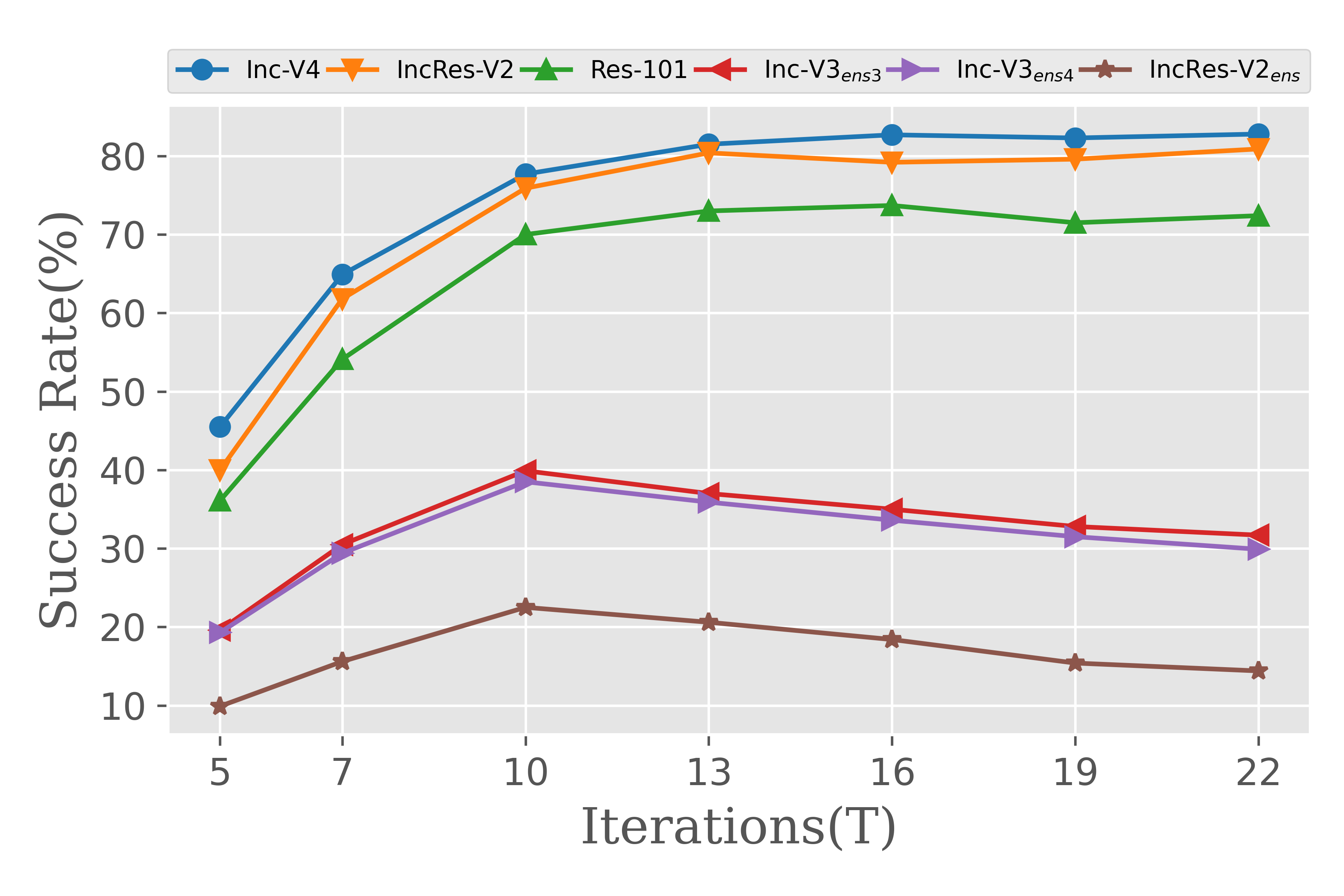}
        \label{fig:5a}
    }
    \subfloat[Inc-V4]{
        \includegraphics[width=0.47\textwidth]{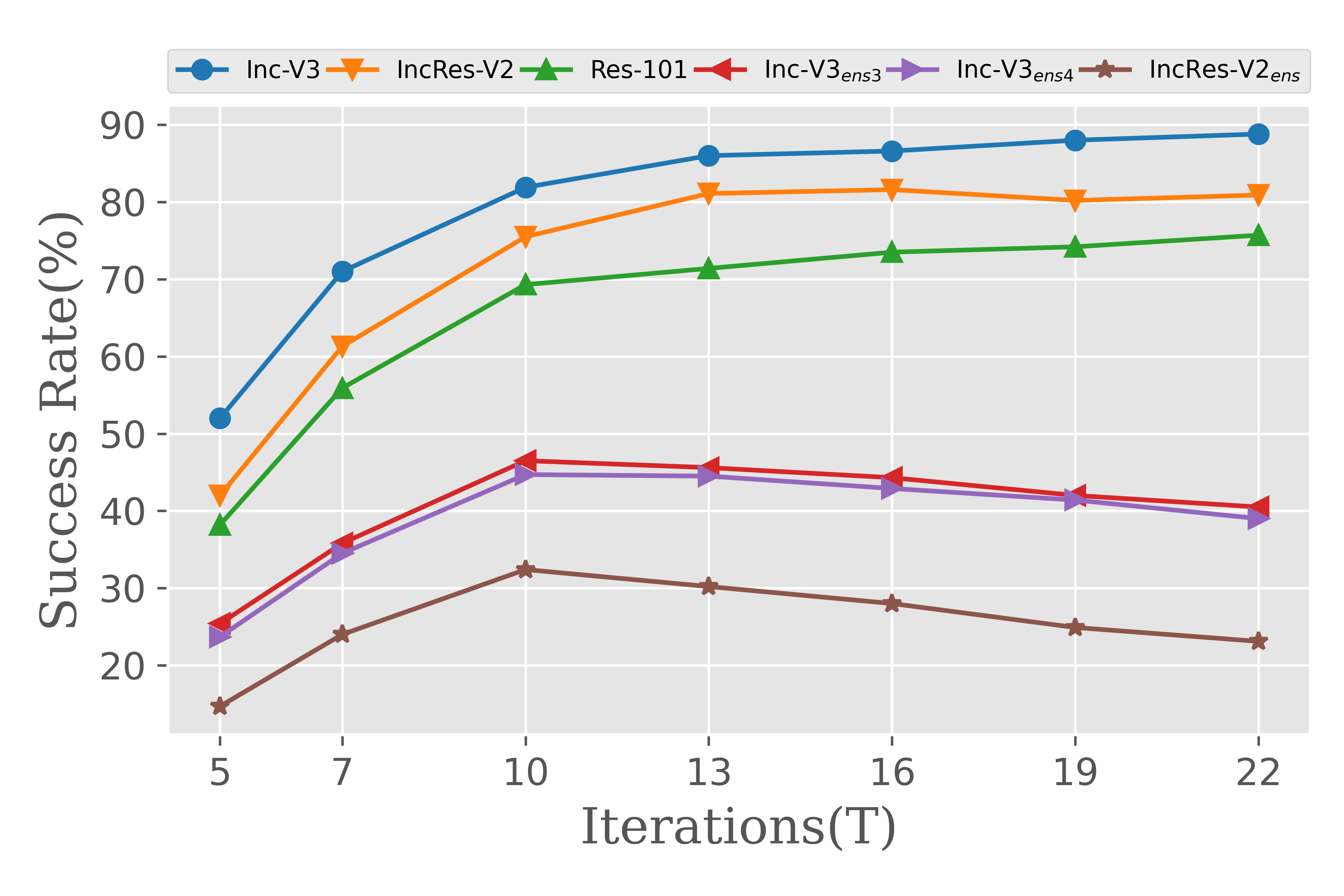}
        \label{fig:5b}
        
    }\\
        \subfloat[IncRes-V2]{
        \includegraphics[width=0.47\textwidth]{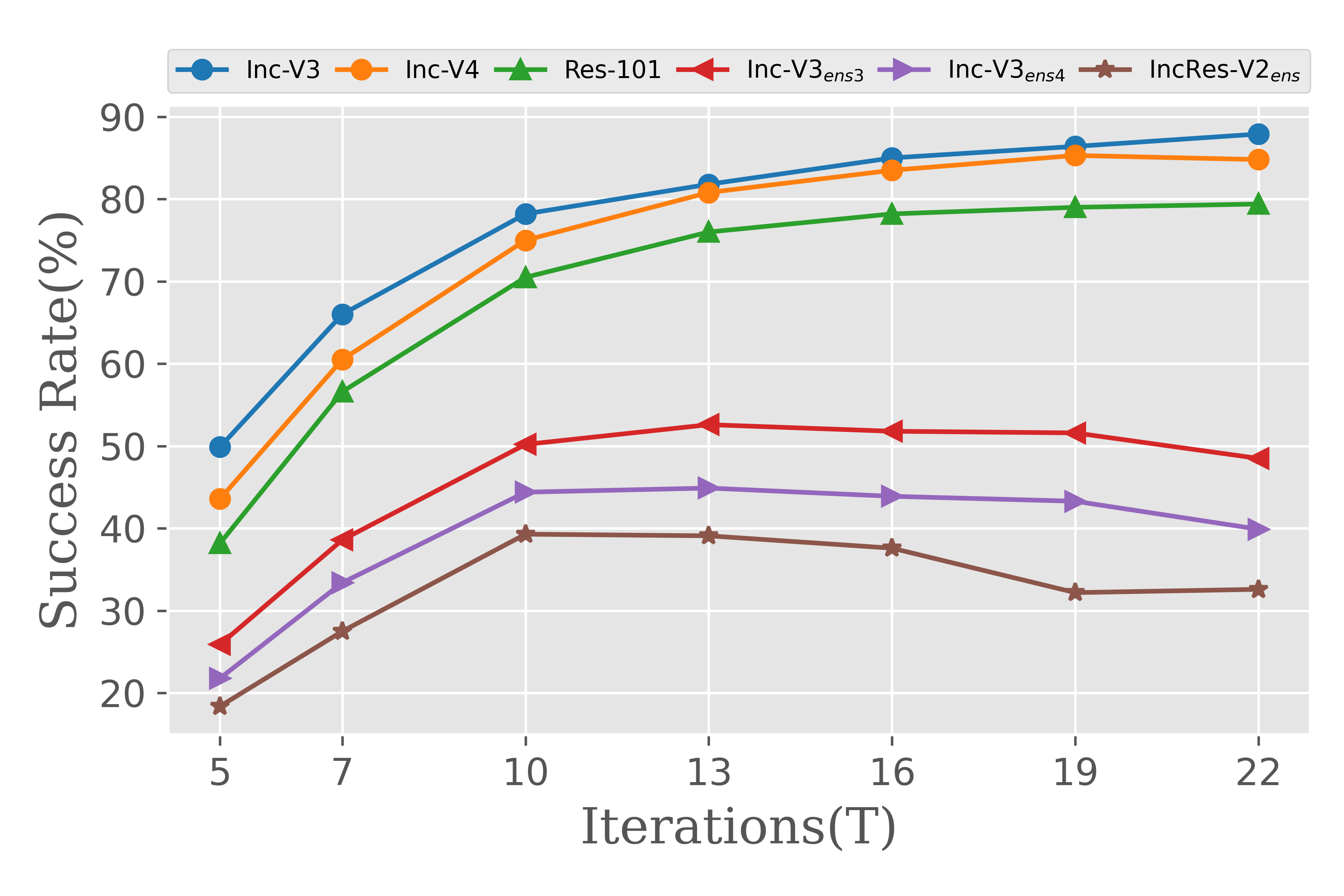}
        \label{fig:5c}
    }
    \subfloat[Res-101]{
        \includegraphics[width=0.47\textwidth]{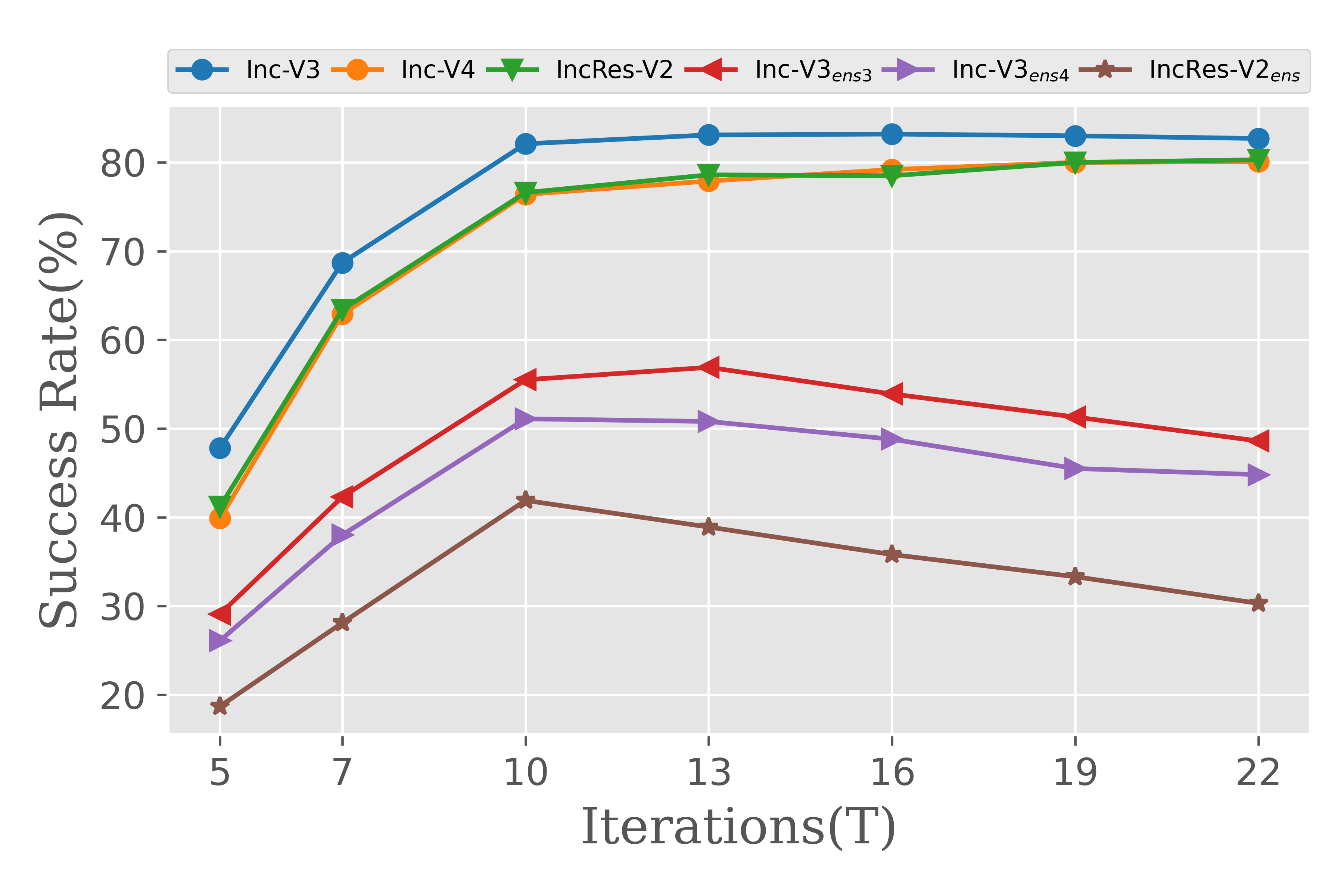}
        \label{fig:5d}
    }
    \caption{The attack success rates ($\%$) of black-box attack against Inc-V3, Inc-V4, IncRes-V2, Res-101, Inc-V$3_{ens3}$, Inc-V$3_{ens4}$ and IncRes-V$2_{ens}$ models when varying $T$ from 5 to 22. The adversarial examples are generated based on Inc-V3 (Fig.~\ref{fig:5a}), Inc-V4 (Fig.~\ref{fig:5b}), IncRes-V2 (Fig.~\ref{fig:5c}) and Res-101 (Fig.~\ref{fig:5d}) models respectively using DA-MI-FGSM attack.}
    \label{fig:5}
\end{figure*}

\textbf{Iterations $T$.} 
We study the impact of iterations $T$ on the transferability of adversarial examples. Similarly, we set sampling times $N$ and standard deviation $\sigma$ to 30 and 0.05 respectively. We fix perturbation size $\epsilon$ to $16$ and step size $\alpha$ to $\frac{16}{10}$. We generate adversarial examples based on normal trained models. Then these adversarial examples are tested on the other models under black-box settings. The total iterations $T$ varies from 5 to 22 and the results are showed in Fig.~\ref{fig:5}.

From Fig.~\ref{fig:5}, we can see that the attack success rates are growing significantly when $T$ is less than 10. However, the attack success rates start to be flattening/slightly growing on the normal trained models and slightly decrease on the adversarial trained models after $T$ is greater than 10. It is worthy to note that the perturbation size reaches the maximum perturbation size because the $\alpha$ is set to $\frac{16}{10}$, which could be the reason why the trends start to be flattening after $T=10$. Besides, we conjecture that the adversarial examples overfit to the white-box model to some extent when $T$ is greater than 10, which decreases its transferability. A similar phenomena can be found on $I-FGSM$ attack which the adoption of multiple iterations decrease its transferability. A possible reason for the steady/slightly increase on the normal trained models when $T>10$ is that the decision boundary of the white-box model is more similar to that of the normal trained models than that of the adversarial trained models.


\begin{figure*}[htb]
    \centering
    \subfloat[Inc-V3]{
        \includegraphics[width=0.47\textwidth]{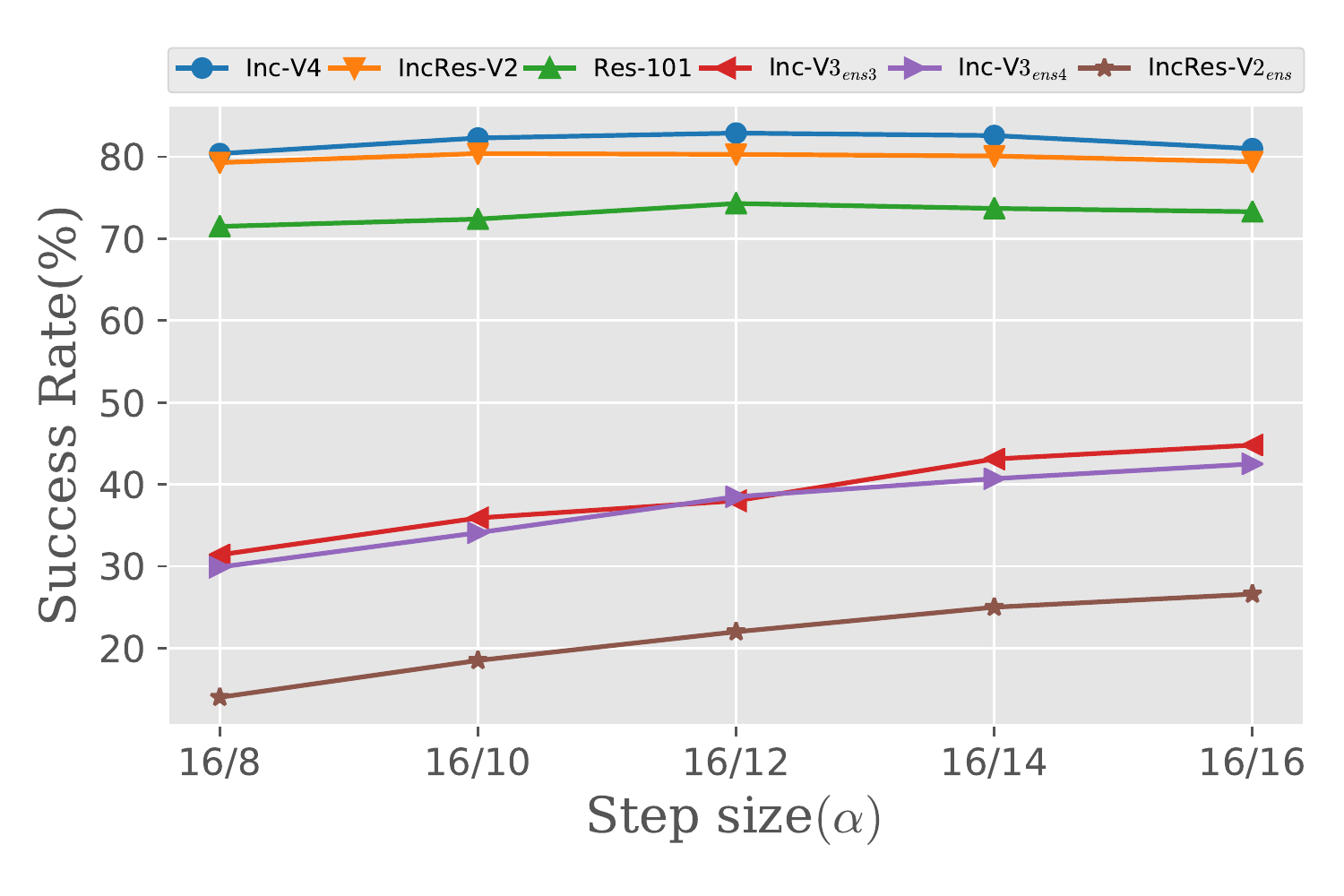}
        \label{fig:6a}
    }
    \subfloat[Inc-V4]{
        \includegraphics[width=0.47\textwidth]{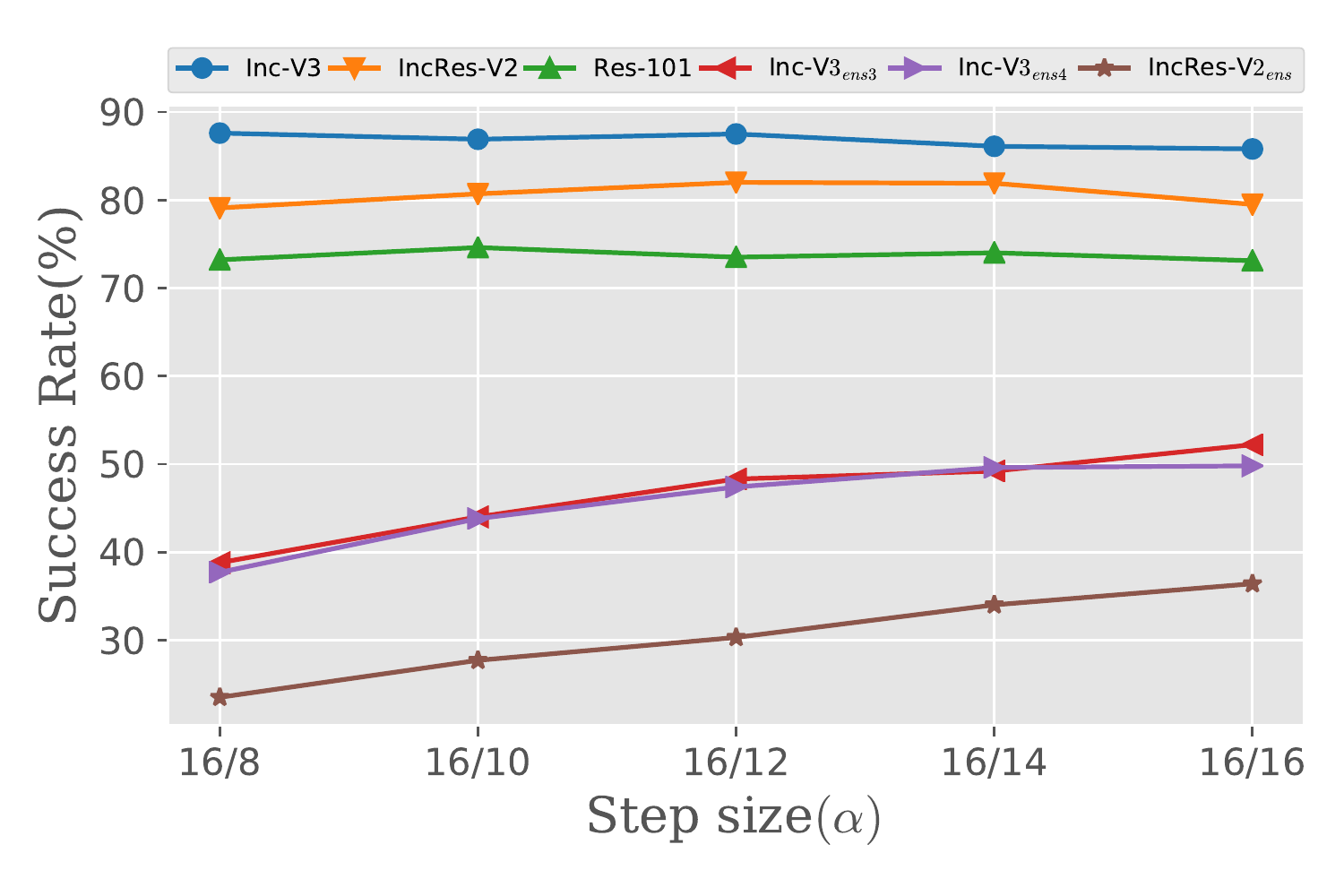}
        \label{fig:6b}
        
    }\\
        \subfloat[IncRes-V2]{
        \includegraphics[width=0.47\textwidth]{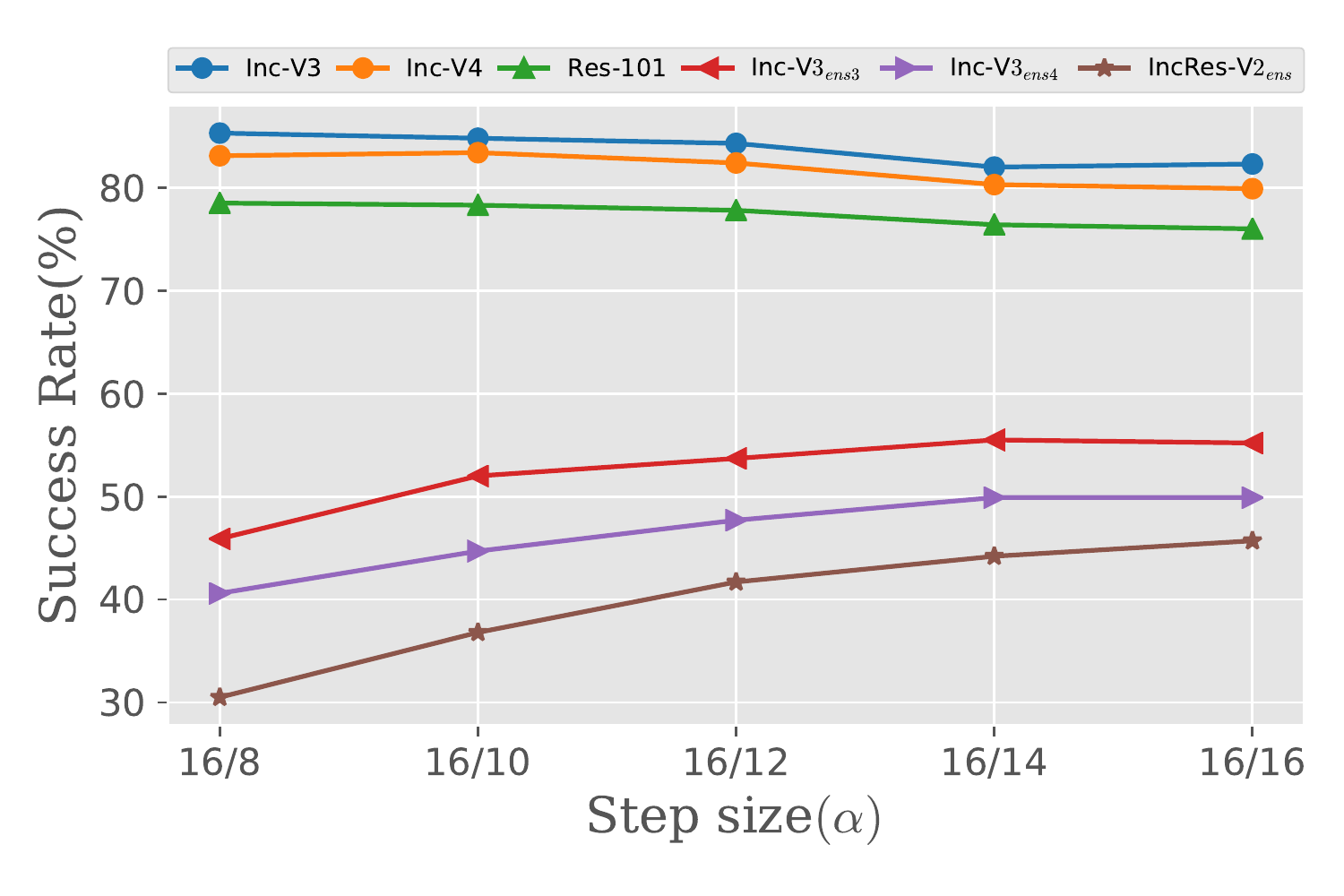}
        \label{fig:6c}
    }
    \subfloat[Res-101]{
        \includegraphics[width=0.47\textwidth]{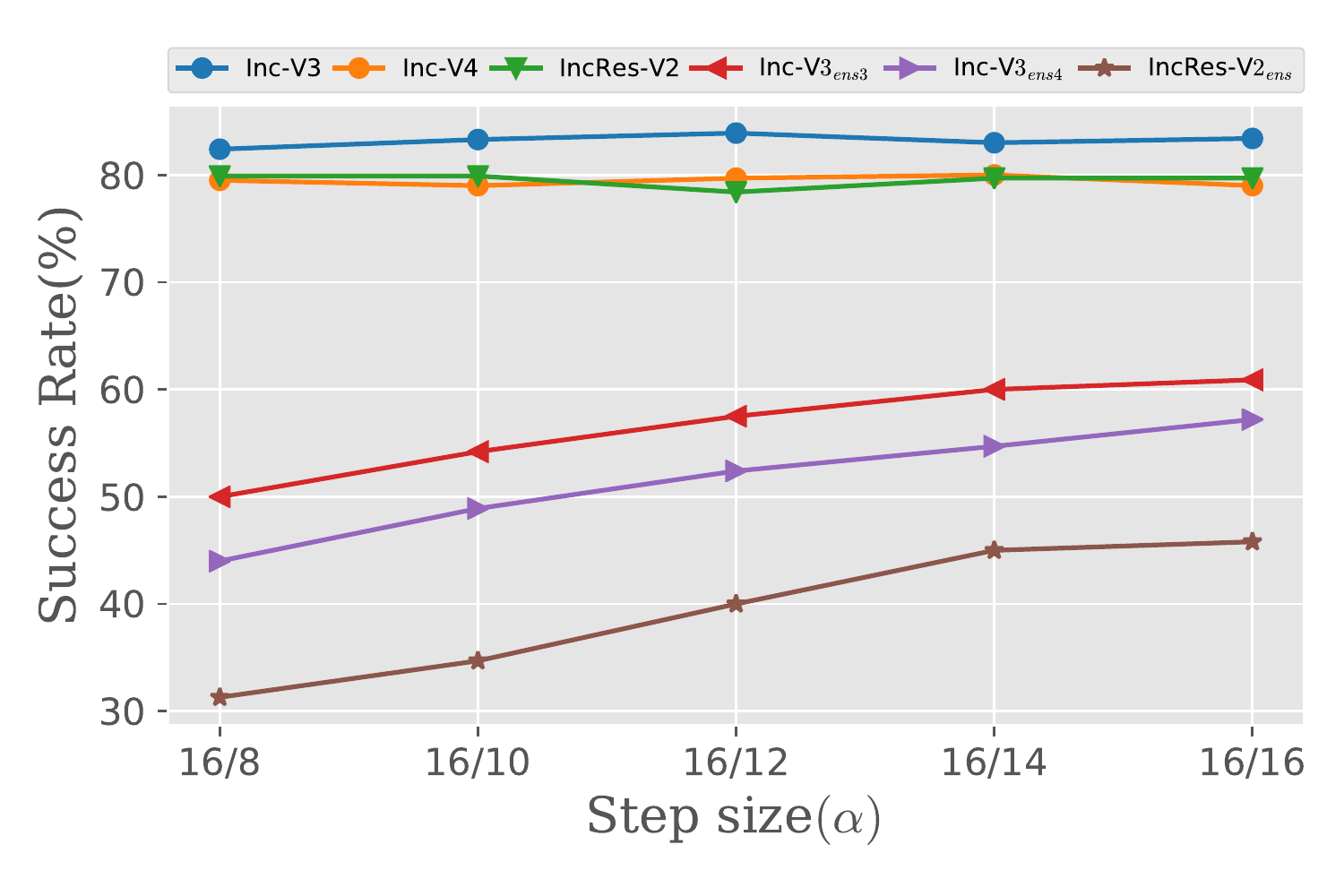}
        \label{fig:6d}
    }
    \caption{The attack success rates ($\%$) of black-box attack against Inc-V3, Inc-V4, IncRes-V2, Res-101, Inc-V$3_{ens3}$, Inc-V$3_{ens4}$ and IncRes-V$2_{ens}$ models when varying $\alpha$ from $\frac{16}{8}$ to $\frac{16}{16}$. The adversarial examples are generated based on Inc-V3 (Fig.~\ref{fig:6a}), Inc-V4 (Fig.~\ref{fig:6b}), IncRes-V2 (Fig.~\ref{fig:6c}) and Res-101 (Fig.~\ref{fig:6d}) models respectively using DA-MI-FGSM attack.}
    \label{fig:6}
\end{figure*}

\textbf{Step size $\alpha$.}
We study the impact of step size $\alpha$ on the transferability of adversarial examples. Similarly, we set sampling times $N$ and standard deviation $\sigma$ to 30 and 0.05 respectively. We fix perturbation size $\epsilon$ to $16$ and iterations $T$ to 16. We generate adversarial examples based on normal trained models. Then we test these adversarial examples on the other models under black-box settings. The step size $\alpha$ varies from $\frac{16}{8}$ to $\frac{16}{16}$ and the results are showed in Fig.~\ref{fig:6}.

From Fig.~\ref{fig:6}, it can be seen that the attack success rates are consistently increasing  with the decrease of $\alpha$ on the adversarial trained models while keep a flat/slightly decreasing trend on normal trained models. The reason for the different trends between normal trained models and adversarial trained models might because the correctly classified samples by normal trained models are very difficult to conduct the transferable attack. To show the evidence for our conjecture, we provide a ratio metric to indicate the percentage of the samples correctly classified by normal models are also correctly classified by the adversarial trained model. We denote $S_{IncV3}=\{\bm{x}\in D^* | f_{\theta}^{IncV3}(\bm{x}) = y\}$ where the mark $IncV3$ denotes the name of the model. The ratio is formulated as follows:
\begin{equation}
    Ratio=\frac{|S_{IncV3}\cup S_{IncV4} \cup S_{IncResV2} \cup S_{Res101} \cap S_{robust}|}{|S_{IncV3}\cup S_{IncV4} \cup S_{IncResV2} \cup S_{Res101}|}
\end{equation}
where the mark $robust$ denotes the surrogate name of adversarial trained models. 

\begin{figure*}[htb]
    \centering
    \subfloat[Inc-V3]{
        \includegraphics[width=0.47\textwidth]{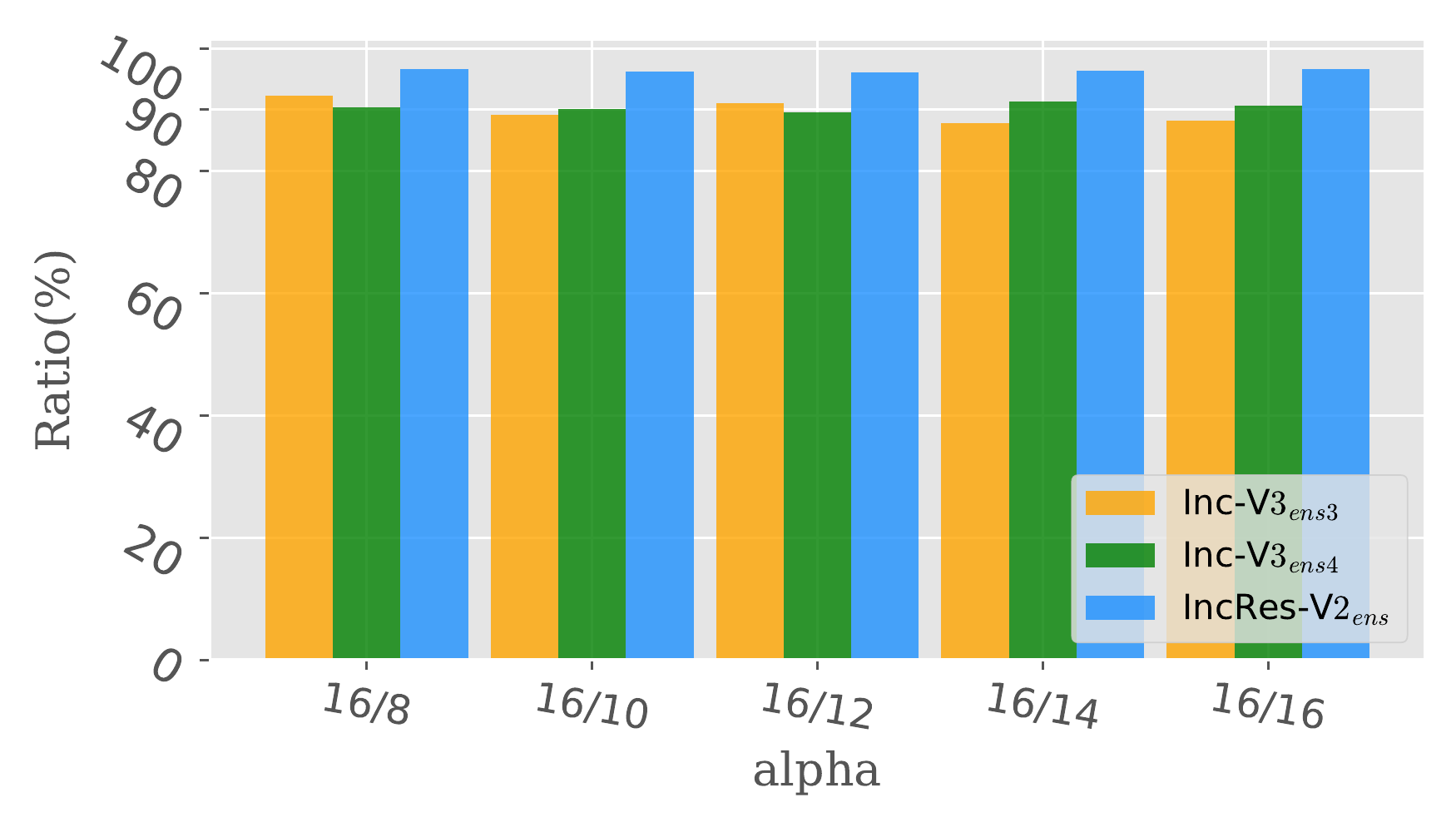}
        \label{fig:ratio:a}
    }
    \subfloat[Inc-V4]{
        \includegraphics[width=0.47\textwidth]{Figures/alpha_ratio_IncV3.pdf}
        \label{fig:ratio:b}
        
    }\\
        \subfloat[IncRes-V2]{
        \includegraphics[width=0.47\textwidth]{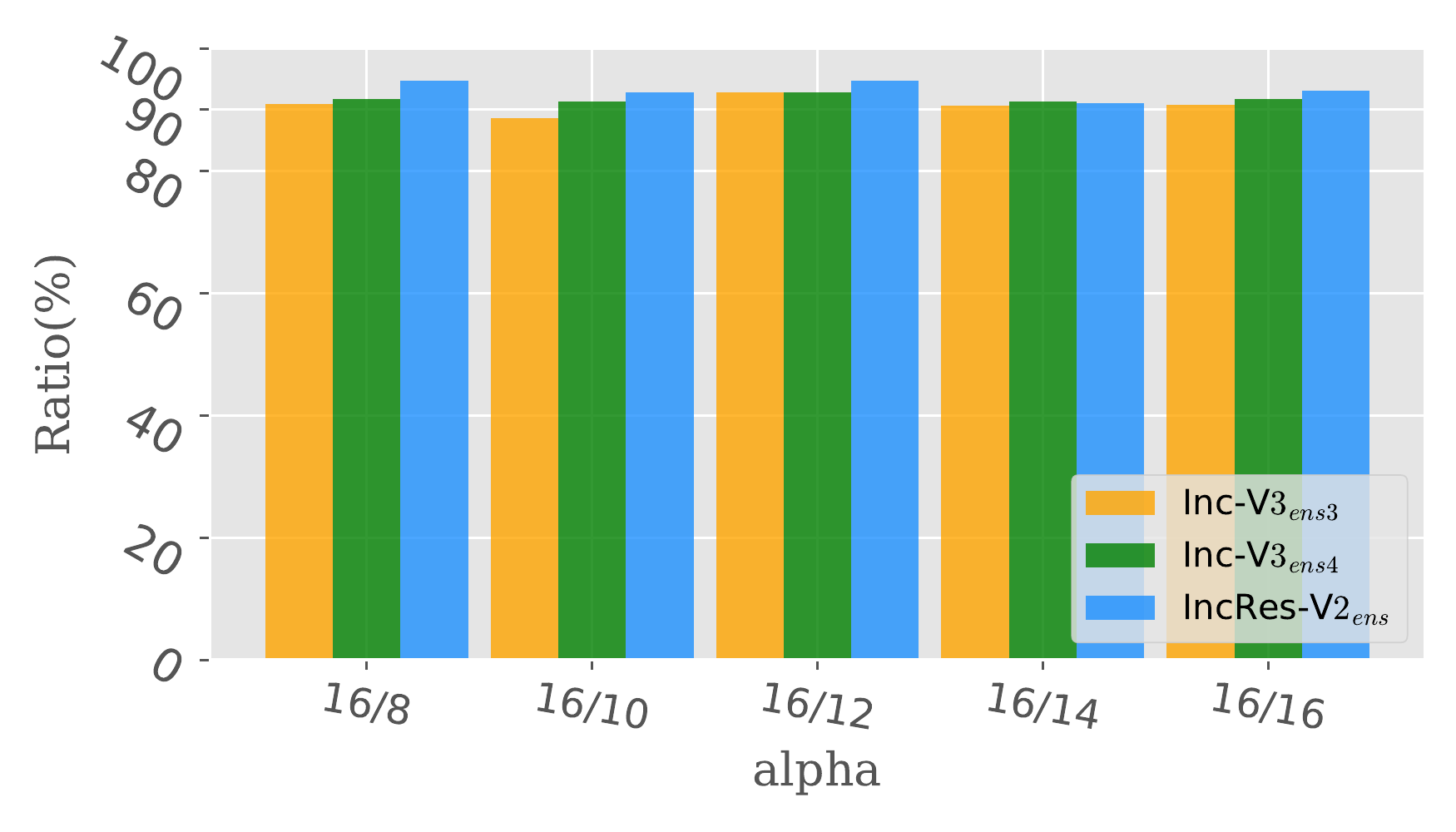}
        \label{fig:ratio:c}
    }
    \subfloat[Res-101]{
        \includegraphics[width=0.47\textwidth]{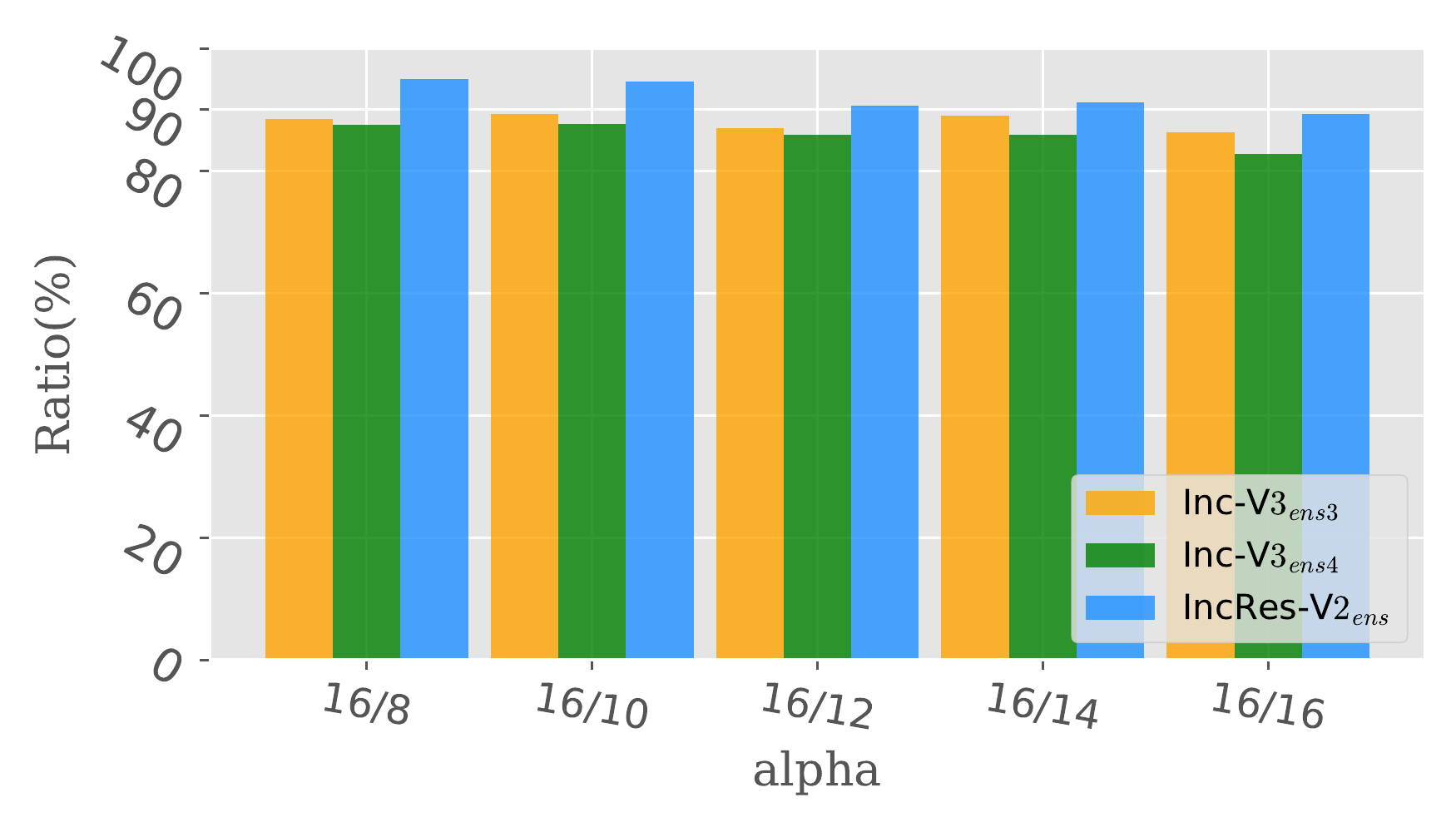}
        \label{fig:ratio:d}
    }
    \caption{The percentage of the samples that are correctly classified by both normal models and the adversarial trained model. Adversarial examples generated by different models are showed in Fig~\ref{fig:ratio:a}, Fig~\ref{fig:ratio:b}, Fig~\ref{fig:ratio:c} and Fig~\ref{fig:ratio:d} respectively.}
    \label{fig:ratio}
\end{figure*}

From Fig.~\ref{fig:ratio}, we can see that around 90\% or more 90\% of the samples that correctly classified by normal trained models are also correctly classified by adversarial trained models. It implies that these samples are difficult to be transferred to attack the black-box models. Therefore, the transferability of these samples improved by reducing $\alpha$ may not be enough to attack the black-box models successfully. 


\section{Connection of the DA-Attack to a Smoothed Classifier}
\label{dis}
Our method mitigates the overfitting problem by aggregating the attack directions of a set of examples around the input $x$, which is different with DIM, TIM and SI-NI-FGSM attacks. 
Essentially, these methods are based on geometric transformations of the inputs, e.g.\ scale and translation. The successful boosting of the performance of combinations of PA-Attack with DIM or TIM (Table~\ref{tab:2}, Table~\ref{tab:3}, Table~\ref{tab:4}) also provides the evidence that our method is orthogonal to these attacks.

For a better understanding of our method, we provide an analyse of connection of DA-Attack to a smoothed classifier. A reasonable assumption is that adversarial examples generated by a non-smoothed classifier are more easily overfitted than that generated by a smoothed classifier.
We take the Gaussian noise smoothed classifier as an example. Formally, given a Gaussian function $g(t)=\frac{1}{\sqrt{2\pi}\sigma} \exp{-\frac{t^2}{2\sigma^2}}$, the Gaussian noise smoothed classifier can be presented as follows:  
\begin{align}
    \Phi(f) (\bm{x}) 
    & =\int_{R^n}g(\bm{y}-\bm{x})f(\bm{y})d\bm{y} \notag \\
    &= \mathbf{E}_{\varepsilon \in \mathcal{N}(0,\sigma^{2} I)}[f(x+\varepsilon)].
    \label{smoothclassifer}
\end{align}
In practice, the Eq.~\ref{smoothclassifer} can be empirical estimated by Monte Carlo sampling. That is, $\Phi(f) (\bm{x})=\frac{1}{N} \sum_{i=1}^{N} f(x+\varepsilon_i), \varepsilon_i \in \mathcal{N}(0,\sigma^{2} I)$. Accordingly, the gradient of $\Phi(f) (\bm{x})$ can be presented as follows:
\begin{align}
    \nabla_{x} \Phi(f) (\bm{x})= \frac{1}{N} \sum_{i=1}^{N} {\nabla_{x} f(x+\varepsilon_i)}, \varepsilon_i \in \mathcal{N}(0,\sigma^{2} I).
    \label{grad_smoothclassifer}
\end{align}
 Comparing Eq.~\ref{grad_smoothclassifer} with Eq.~\ref{eq:2}, it can be observed that when we use the gradient instead of the projected gradient as the update direction, i.e.\ drop the sign function in Eq.~\ref{eq:2}, Eq.~\ref{eq:2} will be equivalent to Eq.~\ref{grad_smoothclassifer}. $\frac{1}{N}$ can be ignored since it will not influence the attack direction. Therefore, our method will be degraded to generate adversarial examples by a smoothed classifier when we use the gradient as the attack direction directly, which also implies that DA-Attack can mitigate the overfitting issue of adversarial examples. 

\begin{figure*}[htb]
    \centering
    \subfloat[Inc-V3]{
        \includegraphics[width=0.47\textwidth]{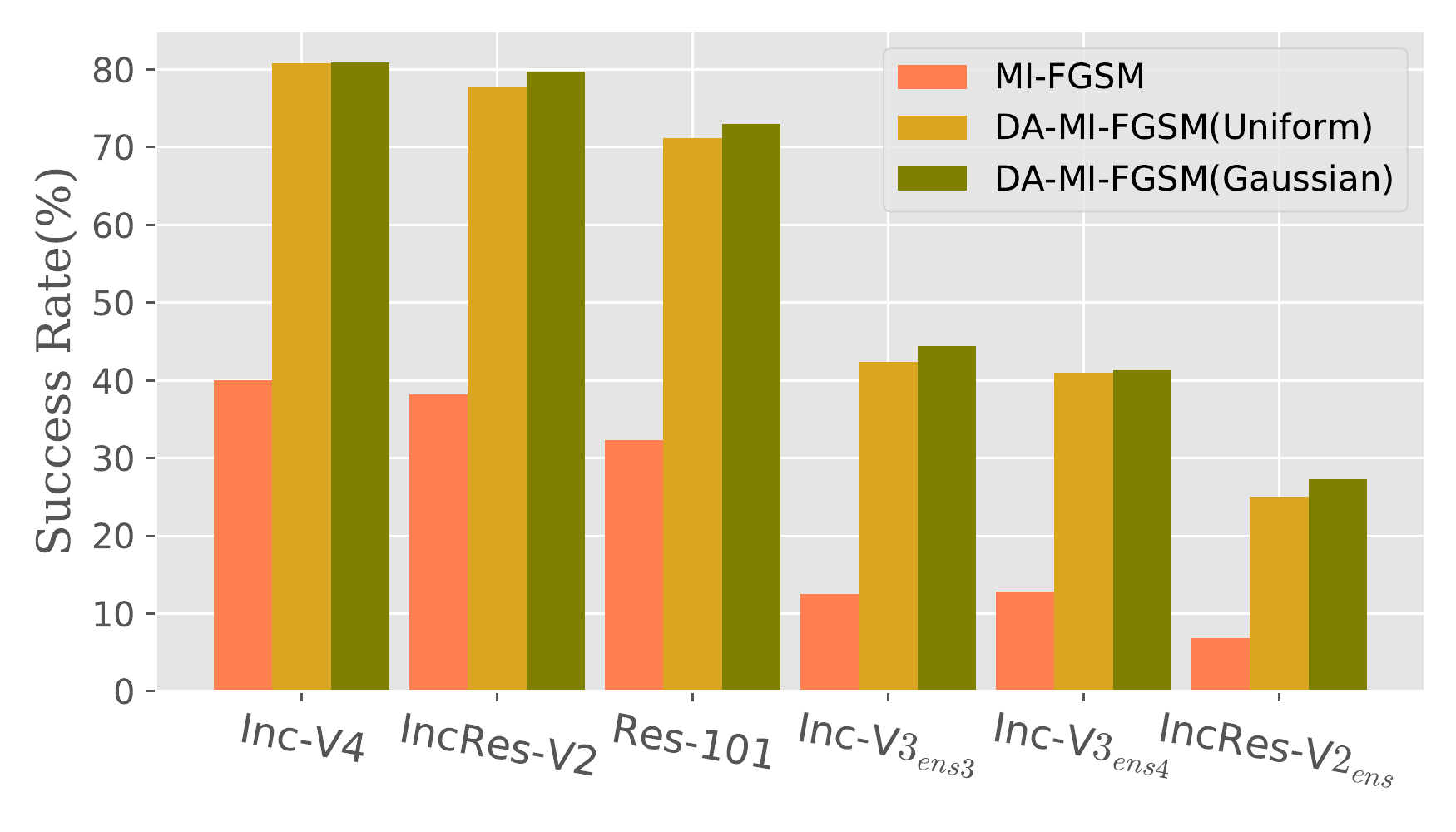}
        \label{fig:7a}
    }
    \subfloat[Inc-V4]{
        \includegraphics[width=0.47\textwidth]{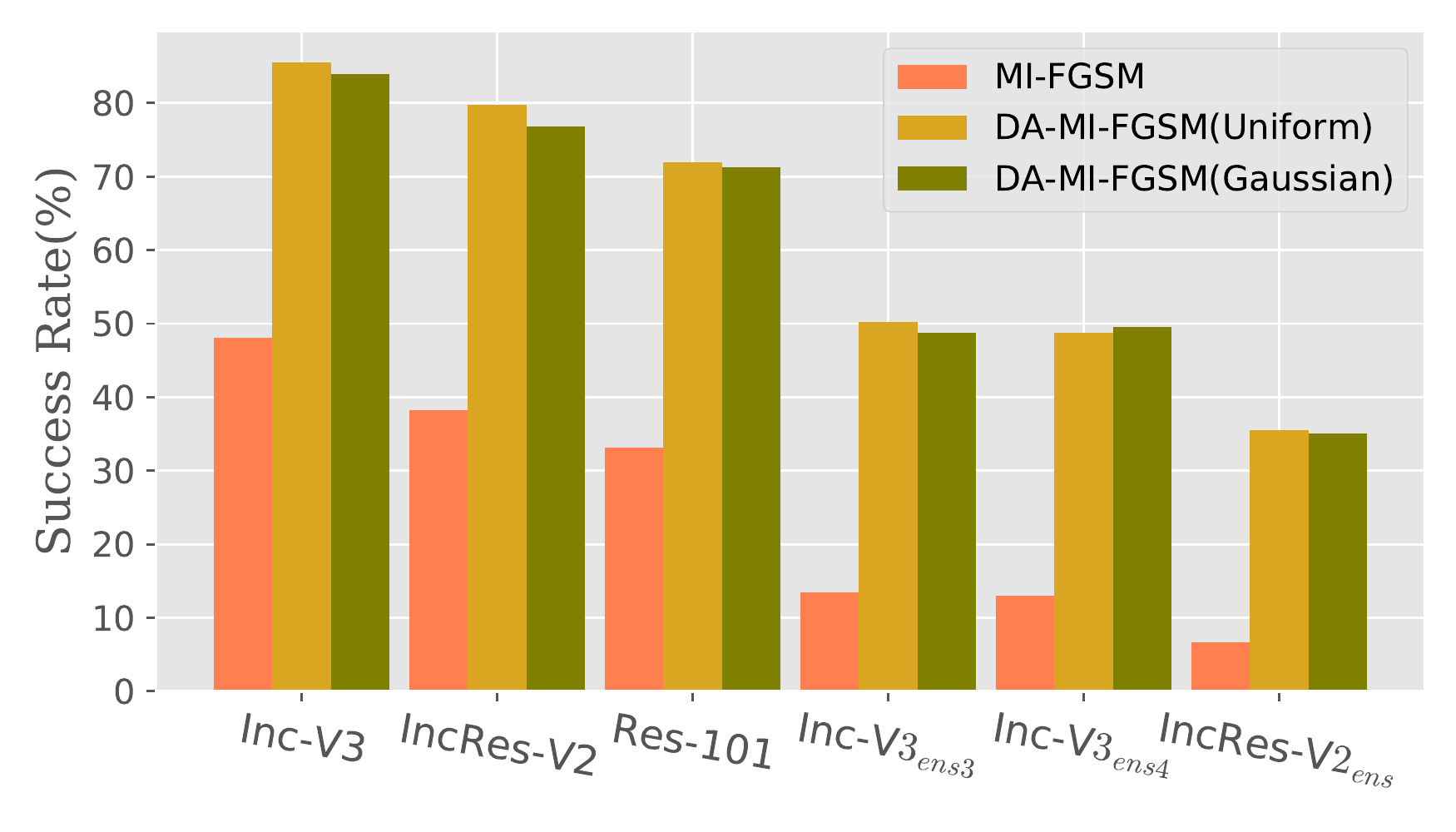}
        \label{fig:7b}
        
    }\\
        \subfloat[IncRes-V2]{
        \includegraphics[width=0.47\textwidth]{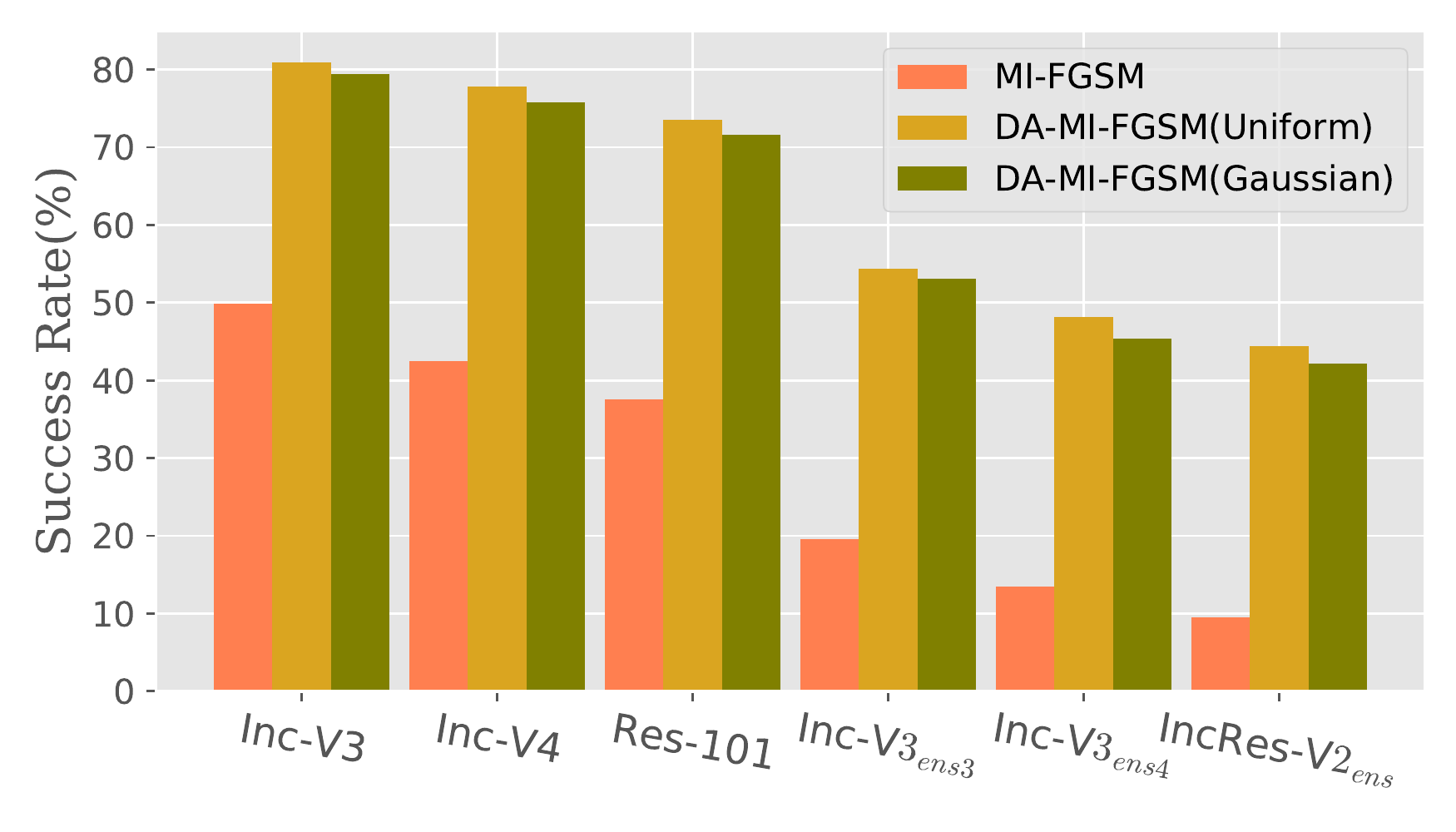}
        \label{fig:7c}
    }
    \subfloat[Res-101]{
        \includegraphics[width=0.47\textwidth]{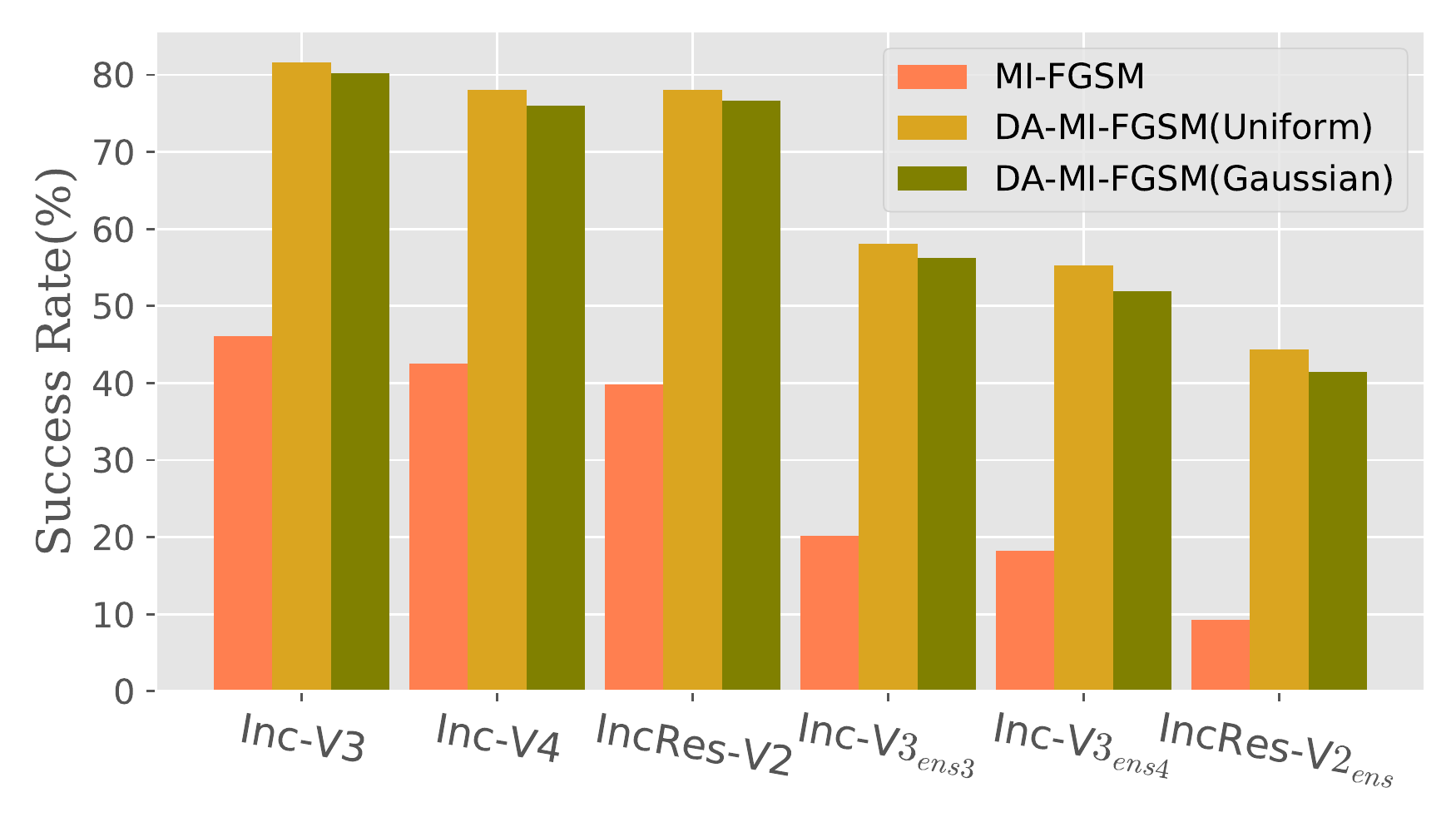}
        \label{fig:7d}
    }
    \caption{The attack success rates ($\%$) of black-box attack against Inc-V3, Inc-V4, IncRes-V2, Res-101, Inc-V$3_{ens3}$, Inc-V$3_{ens4}$ and IncRes-V$2_{ens}$ models. The adversarial examples are generated based on Inc-V3 (Fig.~\ref{fig:7a}), Inc-V4 (Fig.~\ref{fig:7b}), IncRes-V2 (Fig.~\ref{fig:7c}) and Res-101 (Fig.~\ref{fig:7d}) models using DA-MI-FGSM attack with Gaussian noise and Uniform noise respectively.}
    \label{fig:7}
\end{figure*}

Actually, the smoothed classifier also could be smoothed by other noise, e.g.\ Uniform noise, where $g(t)$ is replaced with the uniform distribution function. Similarly,  
Gaussian noise is not the only choice for our DA-Attack. Uniform noise is applicable too. To provide empirical evidence for this, we conduct further experiments by replacing Gaussian noise with Uniform noise (Eq.~\ref{eq:5}) sampled from $\mathbf{U}(-0.08,0.08)$. Other hyper-parameters are set the same as in the preceding experiments (Section~\ref{exp}). The results are shown in Fig.~\ref{fig:7}, from which we observe that DA-Attack with Uniform noise reaches the same performance as when Gaussian noise was added. This experiment illustrates that the choice of the type of perturbations is not the key factor for DA-Attack. 

\section{Conclusion}\label{conclude}
In this paper, we propose to improve the transferability of adversarial examples by aggregating attack directions of a set of examples around the neighborhood of the input. Our proposed DA-Attack makes uses of such aggregated direction. Our extensive experiments on ImageNet with single model attacks and ensemble-based attacks show that our method outperforms state-of-the-art attacks. This result is consistent across all experiments except for the experiments made on IncRes-V2 model. The best averaged attack success rate of our method reaches 94.6\% against three adversarial trained models and 94.8\% against five defense methods under black-box attacks. Our results also reveal current defense models are not safe to transferable adversarial attacks, and therefore, new defense mechanisms are needed. 

We outline several potential approaches for defending against transferable adversarial examples. The essence of existing transferable adversarial examples is that the decision boundaries of the trained models are similar. Therefore, one simple defense approach is to train ensemble models with diversified decision boundaries in order that the decision boundary of each base model is less similar with that of the white-box model.  Another way is  to use transferable adversarial examples as training instances, i.e.\ simply adding them to the training data. This idea is similar to adversarial training. 
The challenge here however is how to generate the on-the-fly transferable adversarial examples efficiently.


 
\bibliographystyle{ACM-Reference-Format}
\bibliography{sample-base}




\end{document}